\documentclass{article}

\usepackage[main, final]{neurips_2026}

\usepackage[utf8]{inputenc} % allow utf-8 input
\usepackage[T1]{fontenc}    % use 8-bit T1 fonts
\usepackage{hyperref}       % hyperlinks
\usepackage{url}            % simple URL typesetting
\usepackage{booktabs}       % professional-quality tables
\usepackage{amsfonts}       % blackboard math symbols
\usepackage{nicefrac}       % compact symbols for 1/2, etc.
\usepackage{microtype}      % microtypography
\usepackage{xcolor}         % colors

\usepackage{amsmath}
\usepackage{graphicx}
\usepackage{caption}
\usepackage{subfig}
\usepackage{wrapfig}
\usepackage{colortbl}  % For cell colors
\definecolor{myred}{RGB}{249, 221, 223} % best tables
\newcommand{\best}[1]{\cellcolor{myred}#1}
\usepackage{multirow}
\usepackage{enumitem}

\colorlet{revcol}{black}
\newcommand{\rev}[1]{\textcolor{revcol}{#1}}
\newcommand{\revcolor}{\color{revcol}}
\usepackage{pgfplots}
\pgfplotsset{compat=1.18}
\usepackage{microtype}
\usepackage{cleveref}
\Crefname{section}{sec.}{secs.}
\Crefname{section}{Sec.}{Secs.}
\Crefname{paragraph}{sec.}{secs.}
\Crefname{paragraph}{Sec.}{Secs.}
\Crefname{table}{tab.}{tabs.}
\Crefname{table}{Tab.}{Tabs.}
\Crefname{figure}{fig.}{figs.}
\Crefname{figure}{Fig.}{Figs.}
\Crefname{equation}{eq.}{eqs.}
\Crefname{equation}{Eq.}{Eqs.}

\title{SyncLight: Single-Edit Multi-View Relighting}

\author{%
  \small David Serrano-Lozano$^1$ \quad
  Anand Bhattad$^2$ \quad
  Luis Herranz$^3$ \\
  \small\bfseries Jean-François Lalonde$^{4, \dagger}$ \quad
  Javier Vazquez-Corral$^{1, \dagger}$ \\[0.5em]
  \small \textsuperscript{1}Computer Vision Center, Universitat Autònoma de Barcelona \quad
  \textsuperscript{2}Johns Hopkins University \\
  \small \textsuperscript{3}Universidad Politécnica de Madrid \quad
  \textsuperscript{4}Université Laval \quad
  \textsuperscript{$\dagger$}Equal advising
}

\begin{document}

\maketitle

\begin{figure}[h]
    \centering
    \vspace{-22pt}
    \includegraphics[width=.98\linewidth]{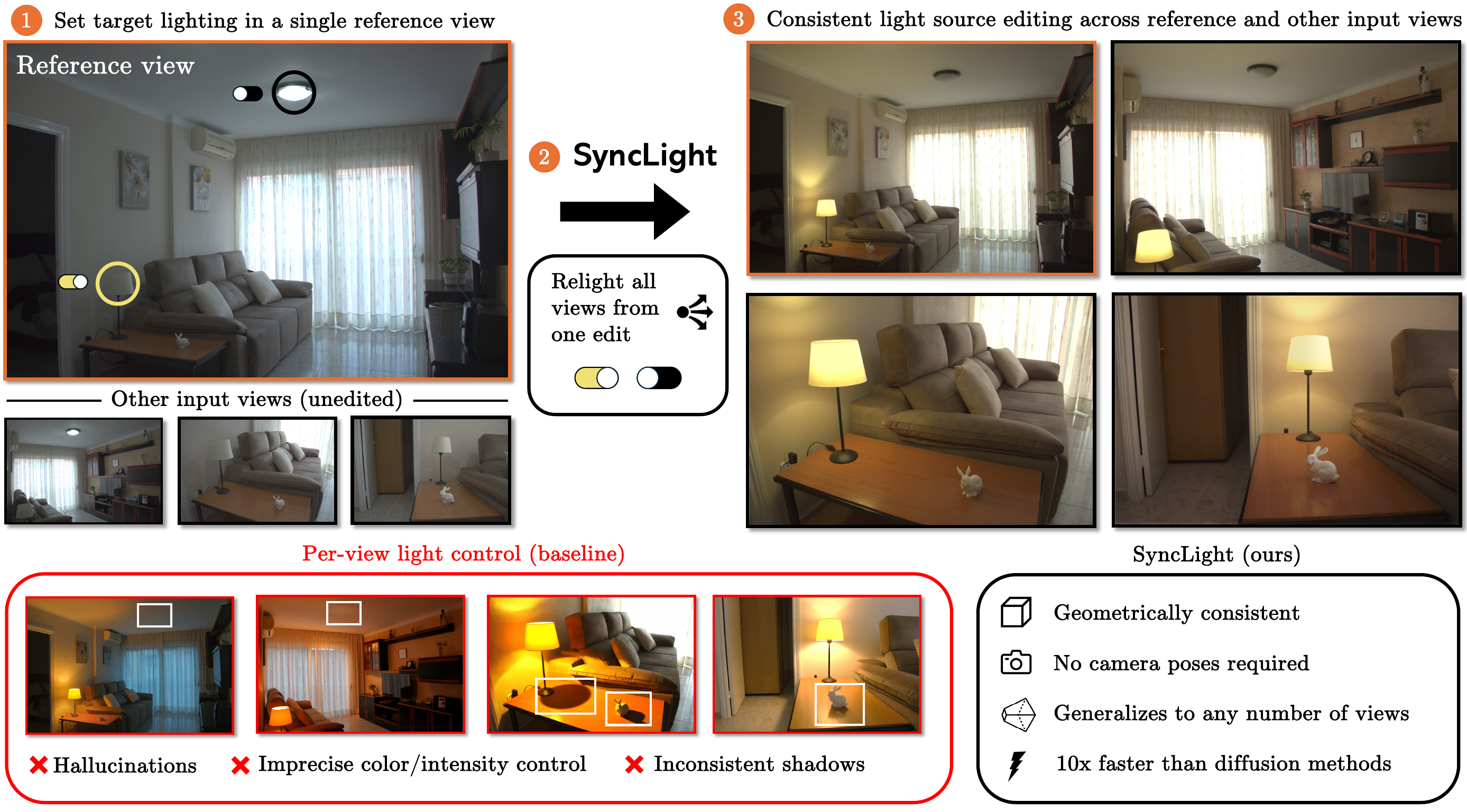}
    \caption{\small Multi-view light editing with SyncLight. Given an uncalibrated multi-view capture of a static scene (1), the user picks a \emph{reference view}, clicks one or more visible light sources (circle markers), and sets their intensity and chromaticity\rev{; the other views never need to be inspected}. SyncLight (2) relights all input views, reference included (3), in a \emph{single} feedforward pass via latent bridge matching. Per-view light control baselines (bottom-left) hallucinate, lack precise control, and are inconsistent across views. SyncLight is geometrically consistent, needs no camera poses, generalizes to any number of views, and is 10$\times$ faster than current models.}    \label{fig:teaser}
    \vspace{-2mm}
\end{figure}

\begin{abstract}
\vspace{-4pt}
We present \textbf{SyncLight}, a method to enable consistent, parametric control over light sources across multiple uncalibrated views of a static scene conditioned on a single view. While single-view relighting has advanced significantly, existing generative approaches struggle to maintain the rigorous lighting consistency essential for multi-camera broadcasts, stereoscopic cinema, and virtual production. SyncLight addresses this by enabling precise control over light intensity and color across a multi-view capture of a scene, conditioned on a single reference edit. Our method leverages a multi-view diffusion transformer trained using a latent bridge matching formulation, achieving high-fidelity relighting of the entire image set in a single inference step. To facilitate training, we introduce a large-scale hybrid dataset comprising diverse synthetic environments---curated from existing sources and newly designed scenes---alongside high-fidelity, real-world multi-view captures under calibrated illumination. Though trained only on image pairs, SyncLight generalizes zero-shot to an arbitrary number of viewpoints, effectively propagating lighting changes across all views, without requiring camera pose information. SyncLight enables practical relighting workflows for multi-view capture systems. \href{https://color.cvc.uab.cat/synclight/}{\textcolor{magenta}{https://color.cvc.uab.cat/synclight/}}
\vspace{-10pt}
\end{abstract}

\section{Introduction}
Controlling illumination after capture in multi-view settings remains an open problem, even as neural rendering~\cite{nerf, gaussian_splatting} has made high-fidelity 3D geometry capture increasingly accessible. This is especially difficult for indoor scenes, where multiple light sources, interreflections, and occlusions create a rich illumination landscape---and where professional workflows like stereoscopic cinema and virtual production demand rigorous consistency, not just plausible appearance.

Relighting is fundamentally an ill-posed problem: from captured images alone, one cannot uniquely decompose appearance into geometry, material, and lighting. State-of-the-art generative methods like LightLab \cite{lightlab} and ScribbleLight \cite{choi2024scribblelight} achieve photorealistic single-image editing, but process each view independently and hallucinate lighting effects that are locally plausible yet globally contradictory: shadows change direction between cameras, specular highlights remain stationary as viewpoints shift. Conversely, inverse rendering and NeRF-based approaches \cite{illuminerf, tensoir} enforce physical consistency through explicit geometry estimation, but require known camera poses, expensive per-scene optimization, and struggle in regions where geometric estimates fail.

To resolve this dichotomy, we introduce \textbf{SyncLight}, a \textit{pose-free} generative framework that brings the flexibility of 2D image editing to multi-view indoor scene relighting. Unlike prior multi-view approaches that depend on calibrated cameras or explicit 3D reconstruction, SyncLight operates directly on uncalibrated image sets: a multi-view diffusion transformer learns to discover cross-view correspondences implicitly through attention, sidestepping both the ill-posed decomposition and the brittle reliance on geometric supervision. A user adjusts light intensity and color on a single reference image, and SyncLight propagates these changes across all synchronized viewpoints in a single forward pass, preserving the complex interplay of shadows, highlights, and indirect illumination.

Realizing this system requires addressing two challenges. First, no existing dataset provides multi-view captures with calibrated lighting ground truth; we introduce a large hybrid dataset combining procedurally generated synthetic environments with high-fidelity real-world captures. Second, prior video-based approaches require per-frame lighting specification, making editing tedious; our multi-view transformer instead propagates a single reference edit jointly to all views, again without any pose information. A key emergent property is \textit{zero-shot scalability}: although trained exclusively on image pairs ($N=2$), our attention mechanism generalizes to dense camera arrays ($N > 2$) at inference without retraining. \Cref{fig:teaser} illustrates this workflow, where users click on the reference image to control intensity and chromaticity, and the edits propagate consistently across all views.

\noindent In summary, we make the following contributions:
\vspace{-5pt}
\begin{itemize}[leftmargin=*,noitemsep]
\item \textbf{Pose-free multi-view relighting:} The first generative method to parametrically relight synchronized, uncalibrated views with strict spatial and photometric consistency in a single forward pass---without any camera poses or explicit 3D reconstruction.
\item \textbf{Zero-shot generalization:} A relighting-aware Multi-View Transformer that learns consistency from image pairs while scaling to arbitrary view counts at inference.
\item \textbf{Efficient one-step inference:} Bridge matching delivers high-fidelity results in a single pass, providing a production-ready alternative to slow optimization baselines.
\item \textbf{SyncLight dataset:} A comprehensive dataset of multi-view indoor scenes under varying illumination, bridging synthetic and real-world domains.
\end{itemize}

\section{Related Work}
The paradigm of relighting has recently shifted from physically-based inverse rendering~\citep{indoor_3d_1, indoor_3d_2, indoor_3d_3, indoor_3d_4, indoor_3d_5, karsch2011rendering, karsch2014automatic, gardner2019deep, gardner2024sky, li2020inverse} to data-driven generative modeling. Rather than explicitly solving the ill-posed decomposition of appearance into geometry, materials, and lighting, modern approaches learn to hallucinate plausible lighting transformations directly in image space~\citep{StyLitGAN, Latent-Intrinsic, LumiNet, lightlab}, leveraging the fact that such decompositions can emerge implicitly within generative models~\citep{bhattad2024stylegan, du2023generative}.

\vspace{-5pt}
\paragraph{Domain-specific methods.}
%Early generative approaches focused on constrained scenarios with strong priors. For portrait relighting, methods leverage light stage captures~\citep{he2024diffrelight, mei2025lux} to model facial appearance under controlled illumination, achieving photorealistic results but limited to frontal face geometry. Object-centric methods~\citep{DilightNet, FlashTex, Neural-Garffer, bharadwaj2024genlit, zhang2024zerocomp, fortier2024spotlight, alzayer2025generativemvr} condition diffusion models on environment/intrinsic maps, spherical harmonics, or shadow masks, enabling compelling material-aware relighting for isolated objects. While effective, %within their narrow scopes, 
%these methods lack the generalization required for complex indoor scenes where multiple interacting light sources and global illumination dominate.

Early generative approaches focused on constrained settings. Portrait relighting methods leverage light stage captures~\citep{he2024diffrelight, mei2025lux} to model facial appearance under controlled illumination, but are limited to frontal faces. Object-centric approaches~\citep{DilightNet, FlashTex, Neural-Garffer, bharadwaj2024genlit, zhang2024zerocomp, fortier2024spotlight, alzayer2025generativemvr} condition diffusion models on environment or intrinsic maps, spherical harmonics, or shadow masks, enabling material-aware relighting for isolated objects. However, these methods do not generalize well to complex indoor scenes with multiple interacting light sources and global illumination.

\vspace{-5pt}
\paragraph{Single-view diffusion for general indoor scenes.}

Recent diffusion-based methods achieve remarkably photorealism by framing relighting as conditional synthesis. IC-Light~\citep{IC-Light} leverages large 2D datasets to hallucinate lighting effects based on text or environment maps, while LumiNet~\citep{LumiNet} and StyLitGAN~\citep{StyLitGAN} explore unsupervised latent control~\citep{Latent-Intrinsic} but suffer from intrinsic ambiguities. Methods such as IntrinsicEdit~\citep{IntrinsicEdit}, IID~\citep{kocsis2024iid}, LightIt~\citep{kocsis2024lightit}, PractiLight~\citep{erel2025practilight}, and RGB$\leftrightarrow$X~\citep{zeng2024rgb} use intrinsic representations for relighting, though they require explicit maps or lack spatial control.

ScribbleLight~\citep{choi2024scribblelight} and LightLab~\citep{lightlab} are state-of-the-art in single-view relighting. 
% conditioning on sparse sketches and depth/segmentation masks, respectively. 
Both process views independently and cannot maintain cross-camera consistency. SyncLight instead uses cross-view attention to propagate a single reference edit across all views in a single forward pass, requiring no per-view masks, depth, or camera poses. Trained only on stereo pairs, it generalizes zero-shot to larger camera arrays at inference. We further replace iterative diffusion (50+ steps) with Latent Bridge Matching for one-step inference (10--50$\times$ speedup).

\vspace{-7pt}
\paragraph{3D-consistent approaches.}
\rev{NeRF and 3DGS-based} methods~\citep{illuminerf, tensoir, alzayer2025generativemvr, poirier2024diffusion, litman2025lightswitch} and inverse rendering approaches enforce consistency through explicit 3D decomposition into geometry, materials, and lighting. While these ensure geometric consistency, they require per-scene optimization and lack flexibility for rapid editing. More recently, \citep{careaga2025physically, lin2023urbanir} combine monocular geometry estimation with physically-based rendering to enable CG-like control over light placement in scenes. While this achieves physical accuracy, it requires explicit mesh reconstruction and does not address multi-view consistency constraints of synchronized camera arrays. SyncLight instead learns consistency priors directly from multi-view data using a generative prior without per-scene 3D reconstruction.

\vspace{-7pt}
\paragraph{Multi-view generation and consistency.}
Recent work has shown that explicit cross-view attention is effective for enforcing geometric consistency. MVDream~\citep{mvdream} introduced multi-view diffusion transformers for text-to-3D generation, demonstrating that shared attention across views yields consistent outputs. We build on this insight for relighting, showing that multi-view transformers trained on stereo pairs generalize zero-shot to larger camera arrays and wide-baseline stereo images never seen during training. Temporal consistency in video---conceptually analogous to spatial cross-view consistency---has been studied in recent video relighting work. RelightVid~\citep{relightvid} and UniRelight~\citep{he2025unirelight} mitigate flickering via temporal modeling and joint denoising. However, these methods operate on sequential frames and do not enforce rigid geometric consistency across viewpoints.

\rev{GR3EN~\citep{xiaoyan2026gr3en} and the concurrent LuxRemix~\citep{ruofan2026luxremix}} also tackle multi-view indoor relighting, but differ considerably from our approach. LuxRemix performs per-view generative OLAT \emph{decomposition} and harmonizes the decomposed lights into a relightable 3DGS representation. GR3EN instead requires \emph{posed} multi-view inputs, reconstructs a radiance field, and distills a video-to-video relighting diffusion model back into the 3D model. Both therefore rely on an explicit per-scene 3D reconstruction, and GR3EN additionally requires known camera poses. SyncLight, in contrast, is pose-free and purely 2D: cross-view consistency emerges from our multi-view transformer, and the entire image set is relit in a single feedforward pass without any per-scene optimization.

\vspace{-7pt}
\paragraph{Flow matching and efficient sampling.}
Our use of Latent Bridge Matching~\citep{chadebec2025lbm} builds on recent advances in flow-based generative models. Flow Matching~\citep{flowmatching} introduced simulation-free training of continuous normalizing flows, enabling faster sampling than standard diffusion. Bridge Matching and Schrödinger Bridge methods~\citep{albergo2025stochastic, shi2023dsbm} further refine these approaches by optimizing transport between arbitrary distributions by adding stochasticity in the trajectory. We adopt this formulation to enable one-step inference essential for interactive multi-camera relighting workflows.

\begin{figure}[t!]
    \centering
    \includegraphics[width=\linewidth]{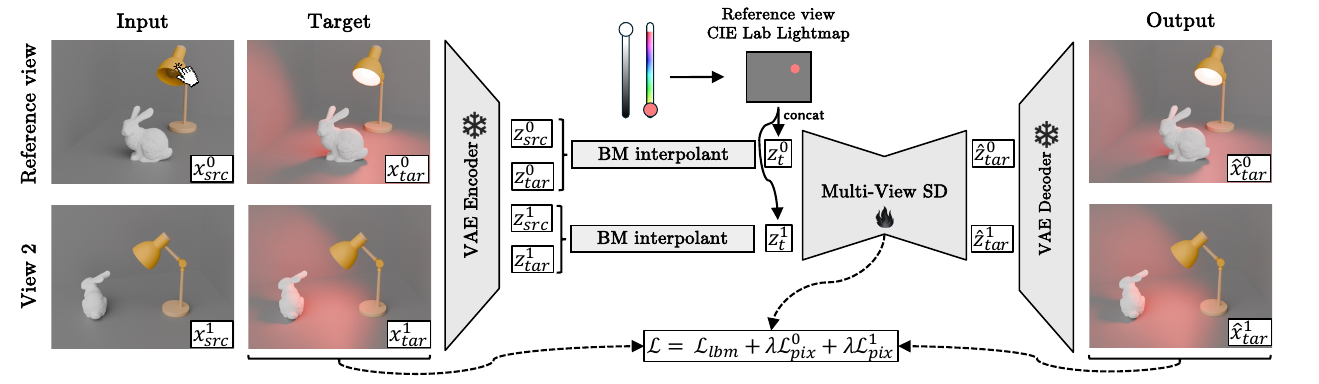}
    \vspace{-15pt}
    \caption{\small SyncLight formulates multi-view relighting as a conditional flow matching problem in latent space. (Left) Input scenes under source ($\mathbf{x}_\text{src}$) and target lighting ground truth ($\mathbf{x}_\text{tar}$) are encoded into latents ($\mathbf{z}_\text{src}, \mathbf{z}_\text{tar}$) via a VAE encoder. This is done for both the ``Reference'' (0) and ``Other'' (1) views. (Middle) During training, we sample a timestep $t$ and construct a Bridge Matching (BM) interpolant $\mathbf{z}_t$. Our backbone, ``Multi-View SD'', is conditioned on a user-specified ``Lightmap'' (encoding color as intensity and chromaticity) derived from the ``Reference view''. To enforce consistency, the backbone processes both views simultaneously. (Right) The model predicts the target latents, which are decoded into relit images ($\hat{\mathbf{x}}_T$). The network is optimized using a hybrid objective $\mathcal{L}$ combining latent flow matching loss ($\mathcal{L}_{\text{lbm}}$) with pixel-level reconstruction losses ($\mathcal{L}_{\text{pix}}$) for each view to ensure high-fidelity, consistent relighting.}
    \label{fig:method}
    \vspace{-5pt}
\end{figure}

\vspace{-1mm}
\section{Method}
\vspace{-1mm}
Our goal is to enable consistent, parametric light-source control across multiple synchronized views by conditioning on parametric lighting edits applied to a single reference view. Unlike prior multi-view diffusion frameworks such as MVDream~\citep{mvdream} or Wonder3D~\citep{long2024wonder3d}, which target image-to-3D generation and require explicit camera pose information, SyncLight is a pose-free, generative framework for editing light sources in synchronized views of an existing scene. Our model implicitly discovers cross-view correspondences through attention, without any geometric supervision, by specifying only the desired light conditions in one reference view. We build upon latent diffusion models~\citep{rombach2022high} and extend them to the multi-view setting through two key contributions: (1) a multi-view transformer with cross-view attention that enforces spatial coherence and consistent propagation of lighting edits, and (2) LBM formulation~\citep{chadebec2025lbm} that enables one-step inference, avoiding the iterative sampling required by standard diffusion models. \Cref{fig:method} provides an overview of our approach.

\vspace{-5pt}
\paragraph{Latent bridge matching.} Traditional latent diffusion models learn to denoise from pure Gaussian noise, which is inefficient for image-to-image translation tasks, where the source image already contains most of the desired scene content. Flow matching~\citep{flowmatching} provides a framework for learning continuous transport between arbitrary distributions. We adopt LBM~\citep{chadebec2025lbm}, which constructs flows between a source latent $\mathbf{z}_{src}$ (encoding an image under source lighting) and a target latent $\mathbf{z}_{tar}$ (encoding the same scene under target lighting). 

We encode an image under source and target lighting $\{\mathbf{x}_\text{src}, \mathbf{x}_\text{tar}\}$ into latent representations using a pretrained VAE encoder, 
$\mathbf{z} = \mathcal{E}(\mathbf{x})$, and define a stochastic interpolation path between them:
\begin{equation}
\mathbf{z}_t = (1 - t)\mathbf{z}_\text{src} + t\mathbf{z}_\text{tar} + \sigma \sqrt{t(1-t)} \boldsymbol{\epsilon} \,,
\end{equation}
where $t \in [0,1]$, $\boldsymbol{\epsilon} \sim \mathcal{N}(0, I)$, and $\sigma$ controls the noise magnitude. Following prior work~\cite{liu2023flow, chadebec2025lbm, tan2025vision}, we use $\sigma = 0.005$, resulting in stochastic trajectories.

The model $\mathbf{v}_\theta$ is trained to predict the velocity field that transports $\mathbf{z}_\text{src}$ to $\mathbf{z}_\text{tar}$. The target velocity is:
\begin{equation}
\mathbf{v} = \mathbf{z}_\text{tar} - \mathbf{z}_\text{src}. 
\end{equation}
The training objective minimizes the mean squared error between predicted and target velocities:
\begin{equation}\label{eq:LBM}
\mathcal{L}_\text{lbm} = \mathbb{E}_{t, \mathbf{z}_\text{src}, \mathbf{z}_\text{tar}}\left[\left|\left| \frac{\mathbf{v}_\theta(\mathbf{z}_t, t, c)}{(1-t)} - \mathbf{v}\right|\right|_2^2\right]\,,
\end{equation}
where $c$ is the conditioning information. We limit $t$ to only 4 equally-spaced timesteps during training.

% This formulation enables efficient one-step inference. 
At inference time, given a source latent $\mathbf{z}_\text{src}$ and a conditioning $c$, the target latent is directly estimated in a \emph{single} step. The decoded output $\hat{\mathbf{x}}_{tar} = \mathcal{D}(\hat{\mathbf{z}}_{tar})$ provides the relit image. Compared to iterative diffusion sampling (50+ steps), this one-step inference is significantly faster while maintaining high visual quality, making it practical for interactive multi-view applications. Observe that at $t=0$, \cref{eq:LBM} collapses to an $L_2$ loss between $\mathbf{v}_\theta$ and the residual $\mathbf{v}$, which might suggest that the LBM formulation reduces to standard regression under one-step inference. However, this is not the case: training across equally-spaced timesteps $t$ forces the network to learn a well-conditioned velocity field along the entire stochastic interpolant, rather than a brittle point-to-point mapping; in line with the observations of~\cite{flowmatching}, this multi-timestep supervision yields more generalizable transport even when inference is truncated to a single step. See Supplementary Material for further discussion.

\vspace{-5pt}
\paragraph{Multi-view latent bridge matching.}

SyncLight takes as input two synchronized, uncalibrated views under source lighting $\{\mathbf{x}_\text{src}^0, \mathbf{x}_\text{src}^1\}$ and a lighting condition $c^0$ defined on the reference view $\mathbf{x}_\text{src}^0$, and predicts both views under target lighting $\{\hat{\mathbf{x}}_\text{tar}^0, \hat{\mathbf{x}}_\text{tar}^1\}$. \rev{Standard diffusion and flow-based} architectures are designed for single-image inputs and do not model cross-view interactions. We therefore use Stable Diffusion XL~\cite{podell2024sdxl} as our velocity field model and adapt it to explicitly reason across multiple views while using its pretrained weights as the model initialization.

Inspired by MVDream~\citep{mvdream}, we replace each transformer block in SDXL with a multi-view transformer block. The key idea is to enable information exchange across $N$ different views through modified self-attention. Given a batch of $N$ view latents, we first process each view's features independently through the initial convolution layers. Before each transformer block, features from all views are concatenated along the token dimension rather than the batch dimension. This allows the self-attention mechanism to attend across views, capturing cross-view geometric and photometric consistency. After attention, features are reshaped back to the original per-view layout. \rev{This rearrangement is essential. In the SDXL UNet, views stacked along the batch dimension never interact in the attention layers, so naively feeding several views to a single-view model does not propagate the edit at all.} Although we train exclusively on pairs ($N=2$), the architecture is agnostic to the number of views due to the token-level attention mechanism. At inference, we observe generalization to arbitrary view counts while maintaining consistency (see \cref{sec:experiments}). Formally, the input to the model $\mathbf{v}_\theta$ is the batch-wise concatenation of all interpolated latent views: $\mathbf{z}_t = \text{concat}[\mathbf{z}_t^0, \cdots, \mathbf{z}_t^{N-1}]$. Despite this architectural inspiration, we note that our setting differs fundamentally: MVDream and related works (e.g., Wonder3D~\citep{long2024wonder3d}) are image-to-3D \emph{generative} models that require camera pose conditioning, whereas SyncLight is a pose-free parametric light source \emph{editing} model operating on synchronous views.

\vspace{-5pt}
\paragraph{Lighting control.}

We represent target lighting conditions via a 4-channel lightmap $L \in [-1, 1]^{H \times W \times 4}$ defined on the reference view $\mathbf{x}_\text{src}^0$. The only constraint on the reference view is that the light sources to be edited must be visible. Users specify locations of light sources to modify (via circular markers), and the lightmap encodes: (1) activation state (1 = on, -1 = off), and (2--4) target color in Lab space. All values default to 0, unless the user specifies the activation and color. We use Lab rather than RGB to enable independent control of brightness (L) and chromaticity (ab). The lightmap is resized to latent resolution and concatenated to each interpolated latent $\mathbf{z}_t^{i}$, yielding 8-channel inputs (4 latent + 4 lightmap). Since the pretrained SDXL backbone expects 4-channel latent inputs, we replace its first convolutional layer with an 8-channel variant whose four additional input channels (corresponding to the lightmap) are zero-initialized. Since all latents must have the same dimensions, we apply the same lightmap to all views---this works well in practice, likely because providing color and intensity hints to other views is helpful, even despite the spatial misalignment. \rev{The model learns that the lightmap lives in reference-view coordinates and associates the marked fixture across views; performance does not degrade with the inter-view angle (\cref{fig:interview_angle}).}

\rev{The interaction is thus one click, plus a color and intensity, on a single reference image per edited light. The user never inspects the other views. SyncLight resolves visibility and occlusion itself, even where the edited source is invisible (\cref{fig:multiview_example}). In contrast, LightLab~\citep{lightlab} and GR3EN~\citep{xiaoyan2026gr3en} need a segmentation mask in every frame (GR3EN also needs camera poses), and LuxRemix~\citep{ruofan2026luxremix} needs Pl\"ucker embeddings and a harmonization stage (\cref{tab:interaction}).}

\vspace{-5pt}
\paragraph{Training objective.}
In addition to the velocity-matching objective ($\mathcal{L}_\text{lbm}$ in \cref{eq:LBM}), we follow prior work~\cite{chadebec2025lbm, yue2023resshift} and impose reconstruction losses on the decoded images:
\begin{equation}
\mathcal{L} = \mathcal{L}_{\text{lbm}} + \lambda \mathcal{L}_{\text{pix}}^0 + \lambda \mathcal{L}_{\text{pix}}^1 \,,
\end{equation}
where $\mathcal{L}_{\text{pix}}^i$ denotes the LPIPS loss~\cite{LPIPS} for view $i$. This perceptual LPIPS penalizes the blur-averaging failure mode, encouraging the network to produce confident, crisp predictions of shadow boundaries.

\section{Dataset}
\label{sec:datasets}

Training a generative model for consistent multi-view relighting requires data satisfying three strict criteria: (1) diverse, high-fidelity indoor geometry; (2) perfectly synchronized multi-view camera poses; and (3) isolated, controllable light sources for supervision. Existing datasets fail to meet these requirements.
%The Multi-Illumination dataset~\citep{murmann2019dataset}, while offering real-world lighting variations, does not contain consistent multi-view captures or visible light sources. The dataset used by LightLab~\citep{lightlab} is single-view only and unavailable.
To address this gap, we introduce the \textbf{SyncLight Dataset}, comprising nearly one million multi-view image pairs under diverse lighting conditions, designed to enable learning of physically consistent relighting across viewpoints.

Following prior work~\citep{hui2018illuminant, aksoy2018dataset, lightlab, heckbert1996synthetic}, we employ a One Light At a Time (OLAT) capture scheme: for each scene, we render or photograph each controllable light source independently, together with an ambient capture containing only non-controllable illumination. All images are captured in linear color space (HDR for synthetic data, camera RAW for real captures) to ensure the photometric linearity required for accurate light composition. We then generate arbitrary lighting conditions by linearly combining these captures with per-light color and intensity coefficients sampled in Lab space. We impose the constraint that the lights being modified must be visible in at least one view, set as the \emph{reference} view, while the second view may or may not have direct visibility of the modified light sources. For our control, we generate a lightmap for the reference view: a mask with colored circles placed at the spatial location of each modified light source, encoding the target color and intensity. %This design forces our model to propagate lighting changes across viewpoints even when the light source itself is occluded in some views, mimicking real-world multi-camera scenarios.

Our dataset comprises three sources with complementary objectives: (i) 365 \emph{Infinigen}~\citep{raistrick2024infinigen} scenes to help the model leverage pretrained scene geometry priors; (ii) 40 \emph{BlenderKit}\footnote{https://www.blenderkit.com/} scenes to provide higher realism, furniture, and light sources; and (iii) \emph{Real} captures from 40 scenes to bridge the domain gap between synthetic and real-world data. In total, the SyncLight Dataset contains nearly one million multi-view training pairs: 920{,}000 from Infinigen, 47{,}000 from BlenderKit, and 18{,}000 from real captures. For testing, we hold out a diverse subset of scenes not seen during training.
Detailed capture protocols, light composition equations, and exemplar images for each split are provided in the Supplementary.

\vspace{-5pt}
\section{Experiments}
\label{sec:experiments}

\vspace{-5pt}
\paragraph{Results and comparisons.}%\rev{We compare against GR3EN~\citep{xiaoyan2026gr3en}, which also addresses multi-view relighting.} No other prior works offer this capability directly, so we \rev{additionally} evaluate baseline methods using various adaptations of state-of-the-art methods. \rev{The code and models of the concurrent LuxRemix~\citep{ruofan2026luxremix} are not available, so we use its synthetic data as an external benchmark instead.}

\begin{table*}[t!]
  \caption{\small Quantitative multi-view image relighting results on our SyncLight test set. Each section of the table indicates which view is used to compute metrics: the ``Reference (Ref.) view'' (where edits are specified), ``Second (Sec.) view'' (another view in the set), and ``Additional (Add.) views'' (all but the reference view). Best results, achieved by SyncLight in all cases, are highlighted. See text for details on baselines. As SyncLight is always run on more than one image; no inference time is reported for the ``Ref. view.'' \rev{VGGTm (\%) measures cross-view consistency against the ground truth (see text); it is undefined for the reference view alone.}}
  \label{tab:quantitative}
  \vspace{-2mm}
  \setlength{\tabcolsep}{1.2pt}
  \footnotesize
  \resizebox{\linewidth}{!}{% camera-ready: VGGTm column added, table scaled to fit (~97%)
  \begin{tabular}{clcccccccccccccc}
    \toprule
    & & \multicolumn{4}{c}{Infinigen} & \multicolumn{4}{c}{BlenderKit} & \multicolumn{5}{c}{Real captures}\\
    \cmidrule(lr){3-6}
    \cmidrule(lr){7-10}
    \cmidrule(lr){11-15}
    &\multicolumn{1}{c}{Method} & PSNR & SSIM & $\Delta E_{00}$ & LPIPS & PSNR & SSIM & $\Delta E_{00}$ & LPIPS & PSNR & SSIM & $\Delta E_{00}$ & LPIPS & \rev{VGGTm} & \rev{Time (s)} \\
    \midrule
    \parbox[t]{2mm}{\multirow{5}{*}{\rotatebox[origin=c]{90}{\footnotesize Ref. view}}} & ScribbleLight & 10.89 & .518 & 30.67 & .382 & 11.02 & .528 & 29.26 & .372 & 11.78 & .483 & 28.36 & .375 & \rev{-} & 58.2 $\pm$ 1.14\\
    &Flux.2-dev & 14.23 & .672 & 20.31 & .338 & 16.71 & .704 & 17.22 & .291 & 18.32 & .757 & 14.28 & .264 & \rev{-} & 55.6 $\pm$ 1.02 \\
    &LightLab* & 29.86 & .941 & 3.08 & .131 & 26.38 & .907 & 4.92 & .147 & 28.42 & .880 & 4.61 & .199 & \rev{-} & 1.17 $\pm$ 0.02 \\
    & GR3EN & 25.92 & .893 & 4.38 & .158 & 21.74 & .803 & 8.93 & .173 & 22.31 & .782 & 9.31 & .238 & \rev{-} & 72.3 $\pm$ 3.51 \\
    & SyncLight & \best{31.32} & \best{.950} & \best{2.47} & \best{.119} & \best{27.16} & \best{.915} & \best{4.02} & \best{.134} & \best{30.34} & \best{.895} & \best{3.73} & \best{.196} & \rev{-} & - \\
    \midrule
    \parbox[t]{2mm}{\multirow{5}{*}{\rotatebox[origin=c]{90}{\footnotesize Sec. view}}}&Scribble+LNet & 13.41 & .498 & 19.67 & .284 & 12.72 & .507 & 23.91 & .306 & 14.48 & .518 & 23.77 & .318 & \rev{50.1} & 69.6 $\pm$ 1.82 \\
    &Flux.2-dev & 12.75 & .617 & 25.82 & .386 & 14.92 & .604 & 20.40 & .469 & 16.35 & .685 & 19.13 & .305 & \rev{52.7} & 111 $\pm$ 2.08 \\
    &LightLab*+LNet & 17.36 & .817 & 7.89 & .184 & 15.92 & .863 & 9.82 & .196 & 20.46 & .815 & 8.46 & .279 & \rev{68.3} & 13.1 $\pm$ 0.68 \\
    & GR3EN & 25.88 & .891 & 4.47 & .163 & 21.43 & .800 & 9.08 & .178 & \rev{22.03} & \rev{.772} & \rev{9.52} & \rev{.244} & \rev{82.7} & \rev{76.4 $\pm$ 3.67} \\
    &SyncLight & \best{31.45} & \best{.949} & \best{2.45} & \best{.119} & \best{28.19} & \best{.925} & \best{3.72} & \best{.126} & \best{30.23} & \best{.897} & \best{3.56} & \best{.197} & \best{\rev{95.8}} & \best{1.58 $\pm$ 0.02}  \\
    \midrule
    \parbox[t]{2mm}{\multirow{5}{*}{\rotatebox[origin=c]{90}{\footnotesize Add. views}}}&Scribble+LNet & 13.36 & .494 & 19.65 & .292 & 12.88 & .509 & 23.47 & .299 & 14.29 & .525 & 23.92 & .320 & \rev{49.8} & 89.7 $\pm$ 1.93 \\ % TODO: VGGTm for Scribble+LNet was not reported in the rebuttal
    &Flux.2-dev & 12.55 & .616 & 26.03 & .388 & 14.90 & .606 & 20.03 & .472 & 16.21 & .678 & 20.42 & .328 & \rev{54.1} & 273 $\pm$ 3.87 \\
    &LightLab*+LNet & 17.22 & .808 & 8.35 & .195 & 16.07 & .871 & 10.43 & .185 & 20.40 & .809 & 8.93 & .287 & \rev{66.8} & 38.5 $\pm$ 0.83\\
    & GR3EN & 25.94 & .894 & 4.32 & .160 & 21.55 & .804 & 8.94 & .174 &  \rev{21.87} & \rev{.769} & \rev{9.72} & \rev{.246} & \rev{82.1} & \rev{91.9 $\pm$ 3.91} \\
    &SyncLight & \best{31.42} & \best{.941} & \best{2.57} & \best{.122} & \best{28.01} & \best{.929} & \best{3.78} & \best{.125} & \best{30.44} & \best{.905} & \best{3.57} & \best{.194} & \best{\rev{94.9}} & \best{2.38 $\pm$ 0.03} \\
  \bottomrule
\end{tabular}}
\vspace{-5mm}
\end{table*}

For single-view relighting of the reference view, we compare against ScribbleLight~\citep{choi2024scribblelight} using Marigold~\citep{ke2023marigold, ke2025marigold} to obtain the albedo and depth, and our own implementation of LightLab \cite{lightlab} (\textbf{LightLab*}). Given that the original model and data are not publicly available, we closely follow the reported design, including a fine-tuned SDXL backbone conditioned on parametric light masks and the original single-view training objective. To ensure a fair comparison, we train our reproduction on the same dataset used for SyncLight, excluding the multi-view component, allowing us to isolate the effect of multi-view modeling. LightLab* serves as our primary single-view baseline and, to the best of our knowledge, represents the current state-of-the-art for parametric single-view relighting. \rev{As LightLab* shares the backbone, data, light conditioning, and training budget with SyncLight, differing only in the multi-view attention and LBM, it acts as a controlled ablation rather than a reproduction of the original model's numbers; its reference-view performance, within 2\,dB of SyncLight on all splits, indicates a competitive implementation.} For relighting the second view, we adopt LumiNet~\citep{LumiNet} (LNet), using the second view as input and the relit version of the reference view as the reference. \rev{We also compare against GR3EN~\citep{xiaoyan2026gr3en}, which addresses multi-view relighting. LuxRemix code and models are not available, so we use its synthetic data as an external benchmark instead (see \cref{tab:luxremix}).}

We additionally compare against Flux.2-dev~\citep{flux2_2025}, a large-scale text-conditioned image editing model. Since it lacks native multi-view support, we implement a two-stage pipeline: we use Qwen-2.5-7B-Instruct~\citep{bai2023qwen} to generate a text prompt describing the desired lighting change from the source and target reference views, apply Flux.2-dev to relight the reference view, and then relight the second view by conditioning Flux.2-dev on the relit reference.

As shown in \cref{tab:quantitative} (``Reference (Ref.) view'' and ``Second (Sec.) view''), our method outperforms previous approaches by a large margin both on the reference and second views. Several key observations emerge from these results. First, SyncLight improves upon Lightlab* on the reference view, indicating that incorporating a second view enhances performance even for single-view relighting. Second, the results for both views in SyncLight are extremely close, clearly indicating the method's ability to consistently match the illumination conditions imposed for the first view with those in the second view. Finally, the average inference time across all images in our dataset demonstrates that our method achieves the most efficient relighting for both views.

\vspace{-5pt}
\paragraph{\rev{Cross-view consistency.}}
\rev{Per-image metrics do not measure cross-view agreement. A model can score well on each view while producing geometrically incompatible views. We therefore introduce VGGTm, the number of high-confidence VGGT~\citep{wang2025vggt} correspondences between the two output views, relative to that between the two ground-truth views (details in \cref{sec:supp_vggtm}). A ratio near 100\% means the outputs are as mutually 3D-consistent as the ground truth. On the Real split, SyncLight retains 95.8\% of the ground-truth correspondences on the second view versus 82.7\% for the best baseline, GR3EN (\cref{tab:quantitative}), confirming genuine cross-view consistency rather than independently plausible per-view outputs.}

\begin{figure}[t!]
    \centering
    \includegraphics[width=.98\linewidth]{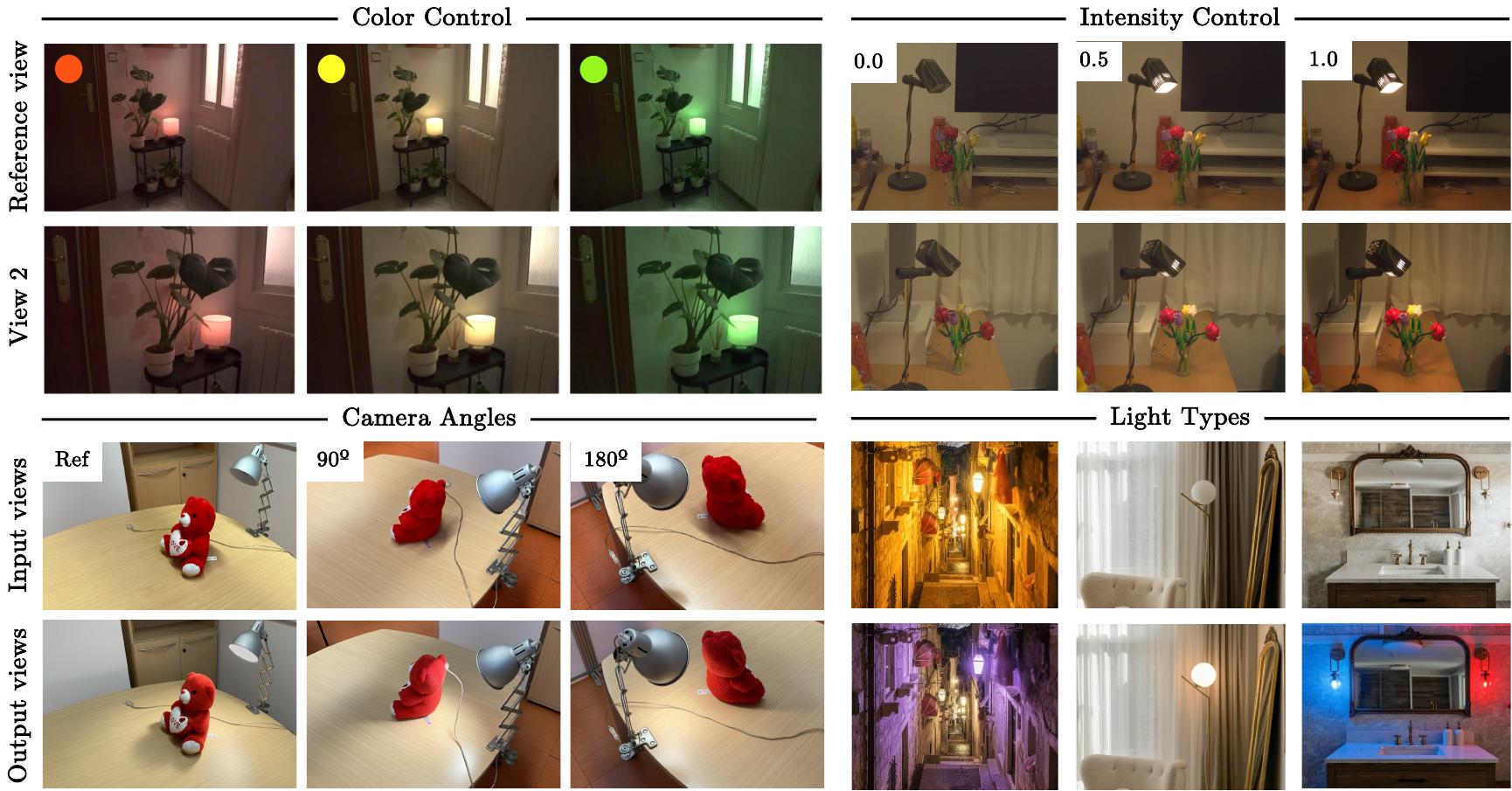}
    \vspace{-1mm}
    \caption{\small Qualitative results showcasing SyncLight's controllability and generalization. \textbf{Top-left:} precise control over light color, varying the chromaticity of a table lamp across three target hues. \textbf{Top-right:} continuous control over light intensity, from off ($0.0$) to fully on ($1.0$). \textbf{Bottom-left:} robustness to large camera angle changes, with input views rotated by up to $180^\circ$ relative to the reference. \textbf{Bottom-right:} generalization across diverse scenes and unseen light types. SyncLight produces consistent relighting across all viewpoints.}
    \vspace{-2mm}
    \label{fig:lightlab-style}
\end{figure}

\begin{figure*}[t!]
    \centering
    \includegraphics[width=\linewidth]{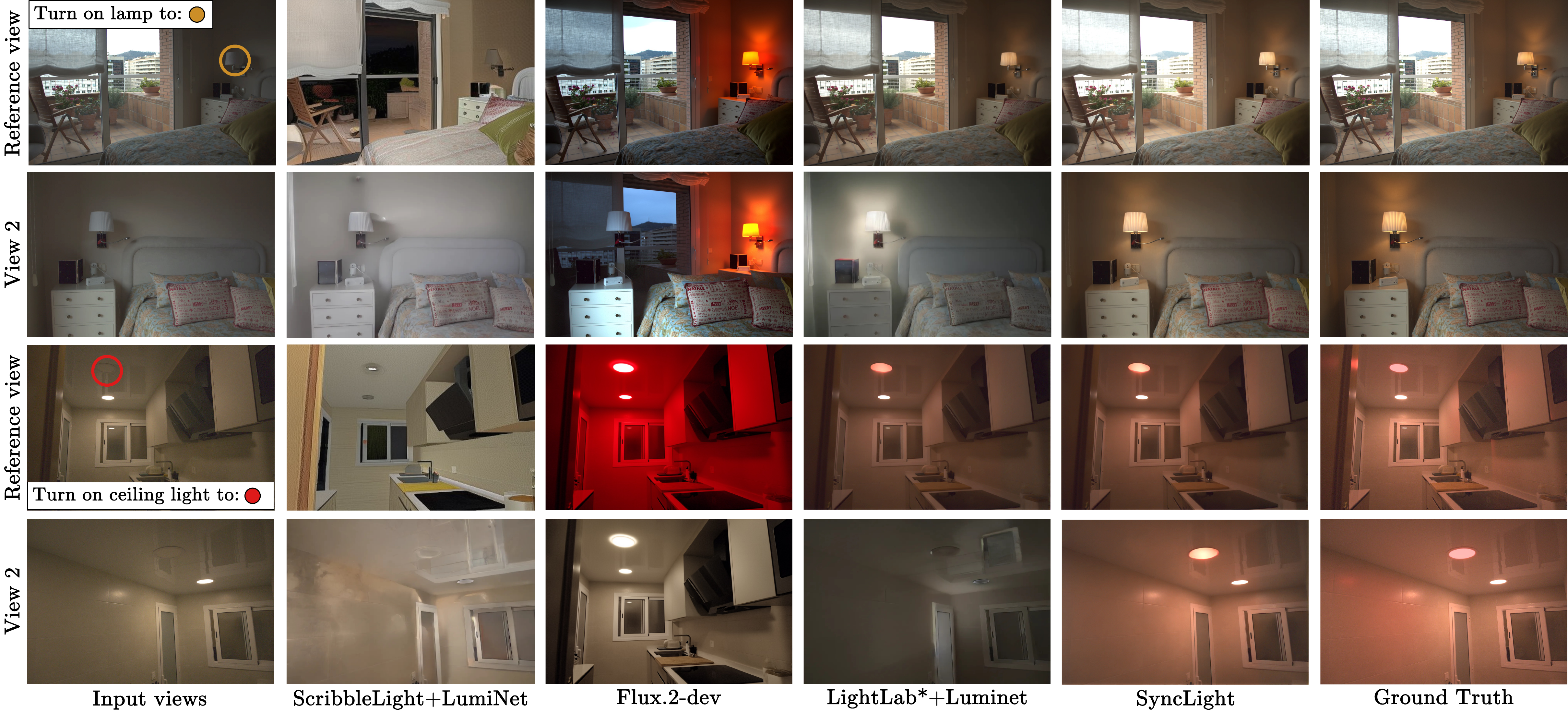}
    \vspace{-6mm}
    \caption{Results in the Real Split from our dataset. SyncLight relights the images correctly. Flux.2-dev struggles with cross-view consistency, producing visually plausible but geometrically inconsistent results (e.g., the window in the second row). Meanwhile, the LightLab*+LumiNet baseline maintains better consistency by explicitly relighting each view, but fails to propagate lighting effects.}
    \label{fig:qualitative_real}
    \vspace{-2mm}
\end{figure*}

\vspace{-5pt}
\paragraph{\rev{Additional benchmark.}}
\rev{To evaluate on data we did not build, we use the OLAT data of the LuxRemix~\citep{ruofan2026luxremix} synthetic dataset to build an external test set of 500 view pairs with relit ground truth, unseen by all methods. SyncLight outperforms all baselines (\cref{tab:luxremix}). It slightly improves on LightLab* on the reference view and, on the second view, gains 5.1\,dB PSNR over GR3EN and 11.1\,dB over LightLab*+LumiNet, the same trend as on our test set.}

\vspace{-5pt}
\paragraph{\rev{Scaling to many views.}}
\rev{To further test the arbitrary-$N$ claim, we render 32 views of each BlenderKit test scene and evaluate $N\in\{2,4,8,16,32\}$ (\cref{tab:scaling}): quality and consistency remain stable. With memory-efficient attention~\citep{dao2022flashattention}, the $NT\times NT$ attention matrix is never materialized, so memory grows only linearly (15.4 to 20.9\,GB); compute grows quadratically but remains reasonable (43.7\,s for 32 views; on a 24\,GB RTX 4090, up to $N=96$ views at 2K in 105\,s). The Real split behaves consistently up to $N=7$ (\cref{tab:scaling_real,fig:scaling_plot}). In this dense setting, which favors video models, GR3EN still lags behind SyncLight while being over 3$\times$ slower and using twice the memory.}
\begin{table}[t!]
  {\revcolor
  \newsavebox{\tabLux}\newsavebox{\tabScale}
  \newcommand{\pmstd}[2]{#1\textsubscript{$\pm$#2}}
  \newcommand{\sbsrow}[2]{\hbox to \linewidth{\hskip 0pt minus 1fil #1\hskip 1em plus 1fill #2\hskip 0pt minus 1fil}}
  \sbox{\tabLux}{\small\setlength{\tabcolsep}{2pt}%
    \begin{tabular}{clcccc}
      \toprule
      & Method & PSNR & SSIM & $\Delta E_{00}$ & VGGTm \\
      \midrule
      \parbox[t]{3mm}{\multirow{4}{*}{\rotatebox[origin=c]{90}{Ref. view}}}
      & Flux.2-dev & 18.67 & .662 & 15.84 & - \\
      & LightLab*  & 25.76 & .800 & 7.18  & - \\
      & GR3EN      & 22.01 & .752 & 10.43 & - \\
      & SyncLight  & \best{25.82} & \best{.803} & \best{7.16} & - \\
      \midrule
      \parbox[t]{3mm}{\multirow{4}{*}{\rotatebox[origin=c]{90}{Sec. view}}}
      & Flux.2-dev     & 15.84 & .656 & 20.83 & 48.2 \\
      & LightLab*+LNet & 14.88 & .682 & 21.92 & 53.8 \\
      & GR3EN          & 20.86 & .738 & 11.92 & 82.1 \\
      & SyncLight      & \best{25.97} & \best{.808} & \best{7.02} & \best{90.6} \\
      \bottomrule
    \end{tabular}}%
  \sbox{\tabScale}{\small\setlength{\tabcolsep}{2pt}%
    \begin{tabular}{clccccc}
      \toprule
      & $N$ & PSNR & SSIM & $\Delta E_{00}$ & Time & VRAM \\
      \midrule
      \multicolumn{2}{l}{GR3EN} & 24.94 & .896 & 8.25 & \pmstd{144}{4.21} & 40.8 \\
      \midrule
      \parbox[t]{3mm}{\multirow{5}{*}{\rotatebox[origin=c]{90}{SyncLight}}}
      & 2  & 28.14 & .941 & 3.64 & \pmstd{1.58}{.02}  & 15.4 \\
      & 4  & 27.92 & .937 & 3.76 & \pmstd{2.17}{.02}  & 15.8 \\
      & 8  & 27.79 & .939 & 3.72 & \pmstd{5.82}{.04}  & 16.9 \\
      & 16 & 28.09 & .940 & 3.73 & \pmstd{17.39}{.05} & 18.3 \\
      & 32 & 27.97 & .938 & 3.77 & \pmstd{43.71}{.07} & 20.9 \\
      \bottomrule
    \end{tabular}}%
  \sbsrow{\begin{minipage}[t]{\wd\tabLux}\vspace{0pt}
      \caption{\small \rev{Additional benchmark built from the LuxRemix~\citep{ruofan2026luxremix} dataset (500 unseen view pairs).}}
      \label{tab:luxremix}
      \par\nointerlineskip\vspace{2pt}\noindent\usebox{\tabLux}
    \end{minipage}}{\begin{minipage}[t]{\wd\tabScale}\vspace{0pt}
      \caption{\small \rev{Scaling with the number of views $N$ (BlenderKit, 32 views per scene). Time (s) and peak VRAM (GB) on an A40. Top: GR3EN with $N=32$.}}
      \label{tab:scaling}
      \par\nointerlineskip\vspace{2pt}\noindent\usebox{\tabScale}
    \end{minipage}}
  }
  \vspace{-1mm}
\end{table}
\vspace{-5pt}
\paragraph{Qualitative results.}
SyncLight enables parametric control over light intensity and chromaticity, and generalizes to diverse scenes and light sources. \Cref{fig:lightlab-style} illustrates four scenarios. \textbf{Color Control} (top-left): SyncLight varies the chromaticity of a table lamp across three target hues at fixed intensity, with color bleed on the wall, plant, and floor propagating consistently to View 2. \textbf{Intensity Control} (top-right): smooth brightness modulation of a desk lamp from $0.0$ to $1.0$, with shadows cast by the flowers and vase sharpening coherently across both views. \textbf{Camera Angles} (bottom-left): given views rotated by approximately $90^\circ$ and $180^\circ$ around the scene, SyncLight produces coherent outputs in which the shadow direction rotates with the camera, suggesting that the model reasons about shadow geometry rather than copying lighting cues from nearby viewpoints. \textbf{Light Types} (bottom-right): generalization to diverse, unseen light sources \rev{(light poles, spherical lights, and vanity-style lights, none of which appear in training)} while preserving the rest of the scene.

\Cref{fig:qualitative_real} shows qualitative results of competing methods and our SyncLight for the Real split of our dataset. As our dataset is color calibrated, we also present the GT on the last column. Results demonstrate our method's key capabilities: (1) Precise light manipulation: SyncLight correctly modifies individual light sources while preserving the remaining illumination, (2) Multi-view consistency: lighting edits propagate coherently across viewpoints, and (3) Indirect effects: SyncLight plausibly synthesizes indirect illumination effects such as wall reflections and color bleeding. In contrast, Flux.2-dev struggles with cross-view consistency, producing visually plausible but geometrically inconsistent results. The LightLab*+LumiNet baseline, while maintaining better consistency by explicitly relighting each view, fails to propagate lighting effects.

\begin{figure}[t!]
  \centering
  \begin{minipage}[t]{0.49\linewidth}
      \centering
    \captionof{figure}{\small Distribution of inter-view angles between reference and second view across our test split (red bars), overlaid with average PSNR per bin (blue curve).}
\label{fig:interview_angle}
    \vspace{-2mm}
    \includegraphics[width=\linewidth]{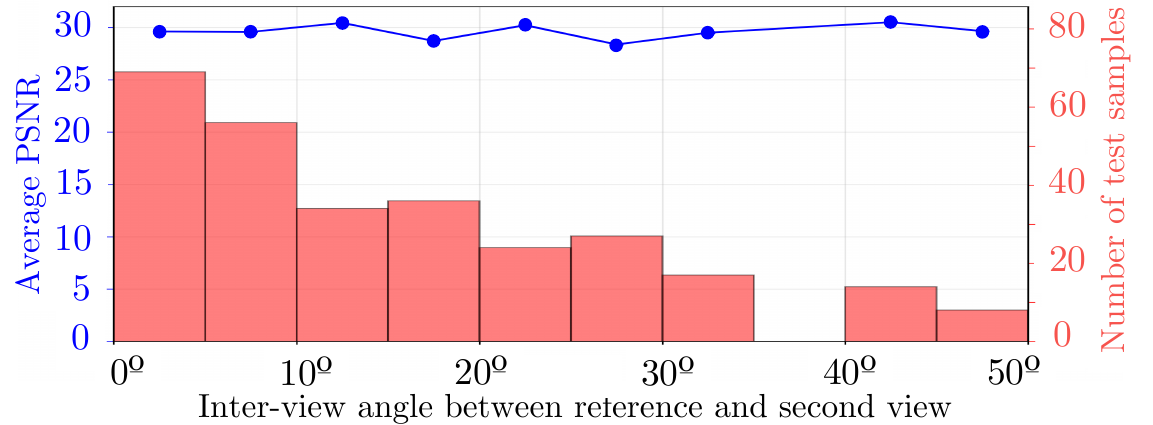}
  \end{minipage}
  \hfill
  \begin{minipage}[t]{0.45\linewidth}
    \centering
    \captionof{table}{\small Ablation studies. Average results on the ``Second view'' of the \emph{Real} split of our dataset.}
    \label{tab:ablations}
    \vspace{-3pt}
    \footnotesize
    \setlength{\tabcolsep}{4pt}
    \begin{tabular}{lcccc}
      \toprule
      & PSNR & SSIM & $\Delta E_{00}$ & LPIPS \\
      \midrule
      w/o \emph{Infinigen}  & 26.91 & .863 & 4.59 & .281 \\
      w/o \emph{BlenderKit} & 27.29 & .872 & 4.42 & .247 \\
      w/o \emph{Real}       & 25.61 & .860 & 4.92 & .238 \\
      \midrule
      w/o MV-SD             & 22.47 & .833 & 5.03 & .236 \\
      \midrule
      SyncLight             & 30.23 & .897 & 3.56 & .197 \\
      \bottomrule
    \end{tabular}
  \end{minipage}  
  \vspace{-12pt}
\end{figure}

\vspace{-5pt}
\paragraph{Inter-view angle performance analysis.} A natural concern with multi-view relighting is whether performance degrades as the inter-view angle between the reference and the second view grows. To evaluate this, we bin the pairs of all test scenes by the angular displacement between the reference and second view, and report the average PSNR within each bin. \Cref{fig:interview_angle} shows the resulting distribution (in red) and the average PSNR within each bin (blue curve). Importantly, the average PSNR remains essentially flat across the entire range, indicating that SyncLight maintains consistent relighting quality even for wide-baseline pairs.
%well beyond the narrow-baseline regime seen at training

\vspace{-5pt}
\paragraph{Ablation studies.} 
\Cref{tab:ablations} presents ablation experiments. Dataset splits: Removing any split degrades performance on the real test split, with \emph{Infinigen} (-3.3 dB in PSNR), \emph{BlenderKit} (-2.9 dB in PSNR), and \emph{Real} (-4.6 dB in PSNR). Multi-view transformer: Replacing multi-view blocks with standard single-view processing causes severe degradation on ``Second view'' (-7.7 dB in PSNR), reducing performance significantly. Without cross-view attention, the model cannot propagate lighting to other views, confirming that feature exchange across views is critical for multi-view consistency.

\vspace{-5pt}
\paragraph{User study.} 
We conducted a 20-participant user study evaluating relighting accuracy, multi-view consistency, and overall quality. Participants compared SyncLight, Flux.2-dev, and LightLab*, the latter provided with an additional lightmap for the second view. SyncLight was preferred in $73.9\%$, $81.1\%$, and $81.7\%$ of cases, respectively, indicating superior perceptual quality and cross-view coherence. See supplementary material for details.

\begin{figure}[t!]
    \centering
    \includegraphics[width=\linewidth]{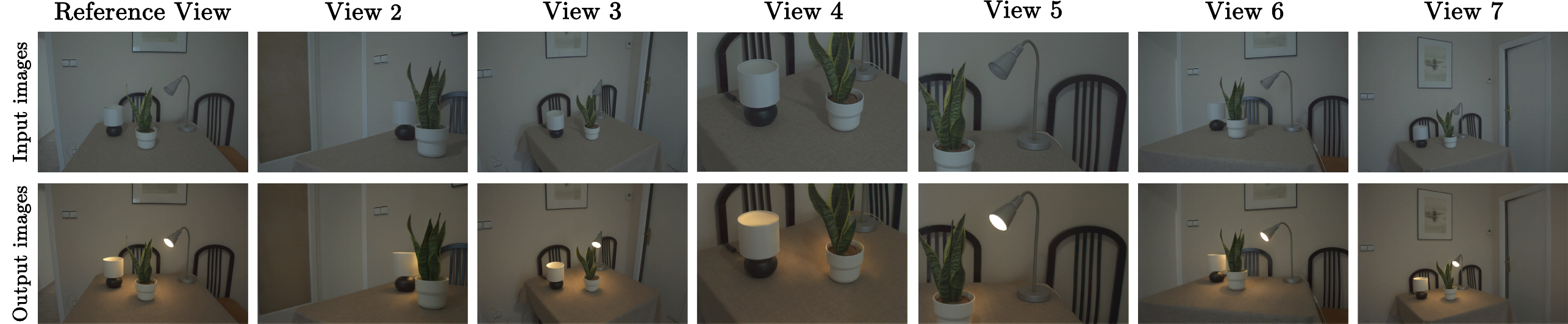}
    \vspace{-5mm}
    \caption{\small Relighting a scene with seven different views. Top row: Input images. Bottom row: SyncLight results. Note how all the views are consistently modified. We want to emphasize views 2 and 4, in which the lamp turned on is not visible, yet the effects in the scene match those of the other views.}
    \vspace{-2mm}
    \label{fig:multiview_example}
\end{figure}

\vspace{-5pt}
\paragraph{Zero-shot multi-view generalization.}
A crucial property of SyncLight is its ability to generalize zero-shot to an arbitrary number of views, despite being trained exclusively on image pairs ($N=2$). This capability stems from our multi-view transformer architecture: because cross-view attention operates at the token level rather than being constrained to a fixed number of views, the learned consistency priors naturally extend to larger camera arrays.
\Cref{fig:multiview_example} presents an example with $N=7$ views, demonstrating consistent relighting across all views despite exceeding the training configuration. Notably, the modified light source is occluded in views 2 and 4, yet the model correctly propagates the lighting changes based on geometric cues from other views. \rev{Moreover}, at 2K resolution, each additional view requires only \rev{$\sim$0.15\,GB} VRAM and 1.23 seconds on an NVIDIA A40.

We quantitatively evaluate this generalization capability in \cref{tab:quantitative} (\emph{Add. views}). For each image pair in our test set, we randomly sample additional views, obtaining an average of 4.38 views per scene. Baseline methods require a separate forward pass for each additional view, always using the first view as a reference. In contrast, SyncLight processes all views in a single forward pass. Our method maintains consistent relighting quality across the reference view, second view, and all additional views, demonstrating robust multi-view consistency beyond the training distribution.

\begin{figure}[t!]
    \centering
    \includegraphics[width=\linewidth]{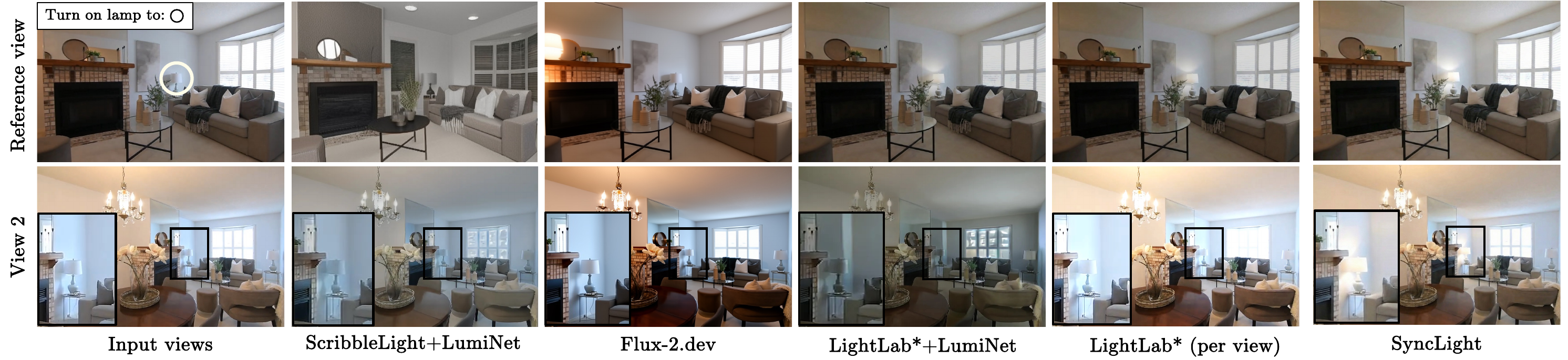}
    \vspace{-5mm}
    \caption{\small Qualitative results on out-of-distribution images from the RealEstate10K dataset~\cite{RealEstate10K}. SyncLight outperforms all the other methods, including a per-view informed version of LightLab*, demonstrating that the joint processing of all views is beneficial even for single-light relighting.}
    \label{fig:qualitative_real_state}
    \vspace{-12pt}
\end{figure}

\vspace{-5pt}
\paragraph{Out-of-distribution scenes.} We additionally evaluate on RealEstate10K~\citep{RealEstate10K} on \cref{fig:qualitative_real_state}, a large-scale collection of indoor video sequences captured with moving cameras across diverse real estate scenes. Although RealEstate10K provides multiple views of the same scene, it lacks ground-truth relighting annotations, and we therefore use it for qualitative evaluation only. \Cref{fig:qualitative_real_state} shows that SyncLight generalizes to these unseen, in-the-wild scenes, producing consistent relighting across viewpoints despite the domain shift from our training data. In the scene the camera displacement is huge, and SyncLight is able to turn on the table lamp correctly. SyncLight is the only method that is able to perform the edits correctly. Note that our method also outperforms a per-view variant of LightLab*, where we relit each view independently, providing an additional lightmap for the second view.

\vspace{-5pt}
   \paragraph{Applications.} Although trained on sparse view pairs, SyncLight also enables video relighting and 3D relighting with radiance fields (appendix~\ref{sec:supp_applications}).

\vspace{-5pt}
\section{Discussion}
We presented SyncLight, a method for consistently controlling light sources across multiple uncalibrated views from a single reference edit, achieving strong qualitative and quantitative performance. \rev{SyncLight assumes (i) a static scene, (ii) synchronized captures, and (iii) that the light sources to be edited are visible in the chosen reference view.} However, our work has limitations that suggest avenues for future research. First, SyncLight relies on a circular region for light conditioning and may struggle with rare or unconventional light sources; in particular, our work does not target windows. Finer-grained segmentation masks, as in LightLab~\cite{lightlab}, could improve generalization to such cases at the cost of increased interaction complexity. \rev{We deliberately avoided mask conditioning (as in LightLab, LuxRemix, and GR3EN) to keep SyncLight self-contained and efficient, without an external segmentation model. SyncLight can fail on very complex fixtures or in scenes with few or misleading visual cues.} Second, leveraging priors from video models~\cite{DiffusionRenderer2025} could enable better generalization to videos, although such approaches may not transfer well to sparse multi-view sets with large baselines, such as those in RealEstate10K. Finally, incorporating 3D priors such as depth or camera pose may further improve performance and broaden generalization. We leave these directions to future work. Despite these limitations, SyncLight operates effectively on wide-baseline stereo setups and demonstrates zero-shot generalization across multiple cameras without relying on 3D information. We are confident that SyncLight marks the beginning of a broader research direction in multi-view, synchronous relighting. \rev{Each ingredient of SyncLight also extends naturally to outdoor scenes, windows and area lights, dynamic multi-camera video (our attention treats its input as a set, so all $N\times T$ latents can be processed jointly), and text-driven agentic interfaces; we outline these extensions in \cref{sec:supp_extensions}.}

\vspace{-9pt}
\noindent \paragraph{Broader impact statement}
Improved relighting models benefit content creation, AR, and VFX by enabling realistic lighting edits. The proposed method is a fundamental research contribution and does not have a direct negative societal impact; however, as with all image generation technologies, downstream applications must prioritize ethical use and guard against potential misuse.

\footnotesize{
\textbf{Acknowledgements.} This work was supported by Grant, PID2024-162555OB-I00 funded by MCIN/AEI/10.13039/ 501100011033 and by ERDF "A way of making Europe", AIA2025-163919-C52 funded by MCIN/AEI/10.13039/501100011033, the grant Catedra ENIA UAB-Cruïlla (TSI-100929-2023-2) from the Ministry of Economic Affairs and Digital Transformation of Spain, and by the Generalitat de Catalunya CERCA Program. JVC also acknowledges the Generalitat de Catalunya CERCA Program, and the 2025 Leonardo Grant for Scientific Research and Cultural Creation from the BBVA Foundation. The BBVA Foundation accepts no responsibility for the opinions, statements and contents included in the project and/or the results thereof, which are entirely the responsibility of the authors. DSL also acknowledges the FPI grant from Spanish Ministry of Science and Innovation (PRE2022-101525). Computing resources were provided by the Digital Research Alliance Canada.}

\small

\bibliographystyle{plain}
\bibliography{main}

%%%%%%%%%%%%%%%%%%%%%%%%%%%%%%%%%%%%%%%%%%%%%%%%%%%%%%%%%%%%
%%% APPENDICES

\clearpage
\appendix
\crefalias{section}{appendix}

\vspace{-5pt}
\section{Dataset details}
\paragraph{Comparison with existing datasets.} Existing datasets fall short of the three criteria required for multi-view relighting (diverse indoor geometry, synchronised multi-view captures, and isolated controllable light sources). The Multi-Illumination dataset~\citep{murmann2019dataset}, while offering real-world lighting variations, does not contain consistent multi-view captures or visible light sources, preventing its use for multi-view propagation. The dataset used by LightLab~\citep{lightlab} is single-view only and is not publicly available. These limitations motivate the construction of the SyncLight dataset.

\vspace{-5pt}
\paragraph{Light composition pipeline.} We first obtain an ambient image $I_\text{amb}$ with all controllable lights turned off, capturing only ambient illumination. Then, we render (or capture) the scene lit by each of the $N$ light sources individually following the OLAT protocol, yielding a set of images $\{I_i\}$ for light $i \in \{1, ..., M\}$. By subtracting the ambient light from each OLAT capture, we obtain the isolated contribution of each light source: $L_i = I_i - I_\text{amb}$. We can then generate arbitrary lighting combinations with different intensities and colors through weighted linear combinations:
\begin{equation}
I_\text{relit} = I_{\text{amb}} + \sum_{i=1}^{M} c_i L_i \,,
\end{equation}
where $c_i \in \mathbb{R}^3$ specifies the desired color (and intensity) of light $i$. In practice, $c_i$ is originally defined in Lab space as to more intuitively separate luminance from chromaticity and then converted to RGB before applying the combination. Finally, we apply tone mapping to convert the linear result to a non-linear 8-bit image. We use Reinhard method with a $\gamma=2.2$.

\vspace{-5pt}
\paragraph{Dataset sources.} Our dataset presents images from three different sources: Infinigen, BlenderKit, and Real captures. Infinigen images enable the model to leverage pretrained scene‑geometry priors across views. BlenderKit images contribute increased realism, including furniture and light sources. Finally, Real capture images help reduce the domain gap between synthetic and real‑world data.

\vspace{-5pt}
\paragraph{Infinigen.} We utilize Infinigen~\cite{raistrick2024infinigen}, a procedural generator of photorealistic 3D Blender scenes, to create 365 diverse indoor environments across five room types: bathrooms, bedrooms, living rooms, dining rooms, and kitchens. Each scene exhibits high variation in geometry, materials, and assets. We procedurally place 5 camera groups, each comprising 5 cameras, pointing at different parts of the scene. We ensure at least 25\% spatial overlap between all viewpoints within the same group to maintain multi-view consistency constraints. For each scene, we identify all emissive objects (e.g., lamps and ceiling lights) and render them following the OLAT protocol. Each view is rendered in HDR with physically-based rendering, producing a total of 920,000 pairs of different lighting combinations. 

\vspace{-5pt}
\paragraph{BlenderKit.} To increase realism beyond procedurally-generated scenes, we curate 40 high-quality Blender scenes from BlenderKit\footnote{https://www.blenderkit.com/}, a repository of artist-created 3D assets and scenes. These scenes feature more realistic furniture arrangements, diverse architectural styles, and complex lighting fixtures. To integrate these scenes into our pipeline, we perform several preprocessing steps: (i) we replace all emissive materials with transparent materials and explicit Light objects to enable precise OLAT control; (ii) we manually position 3--6 cameras per scene following the same multi-view overlap constraints as \emph{Infinigen}; (iii) we remove unrealistic or low-quality assets; and (iv) we augment scenes with additional BlenderKit assets including diverse lamps, light fixtures, and decorative objects to enhance scene complexity and realism. Each view is rendered in HDR format, similarly to Infinigen, producing 47,000 pairs with diverse lighting combinations.

\vspace{-5pt}
\paragraph{Real captures.} To bridge the gap between synthetic and real-world imagery, we capture 320 RAW photographs from 40 indoor scenes. We use a DSLR camera (Canon EOS 1300D) and a mobile device (iPhone 13), fixed on a tripod and activated via a remote trigger. For each viewpoint, we fix the camera parameters (ISO, shutter speed, and aperture for the DSLR; ISO and shutter speed for the mobile device) to preserve photometric consistency across OLAT captures. Exposure settings are selected to minimize clipping under active lighting while keeping the ambient capture sufficiently bright to avoid excessive noise. We reposition the tripod to acquire multiple viewpoints per scene and capture each scene following the same multi-view OLAT protocol as in the synthetic splits. We then manually annotate light source locations in the images and apply the same light composition pipeline to generate new lighting conditions. This yields 18,000 real-image pairs with diverse lighting configurations. 

\vspace{-5pt}
\paragraph{Test split.} For testing, we hold out a diverse subset of scenes not seen during training: 10 from Infinigen, 3 from BlenderKit, and 6 from Real captures. From these unseen scenes, we randomly sample a fixed test set of 135, 45, and 105 pairs, respectively. This held-out split is used for all quantitative evaluations reported in the main paper.

\begin{figure*}[t!]
    \centering
    \subfloat[\textit{Infinigen} split\label{fig:qualitative_infinigen}]
    {\includegraphics[width=\linewidth]{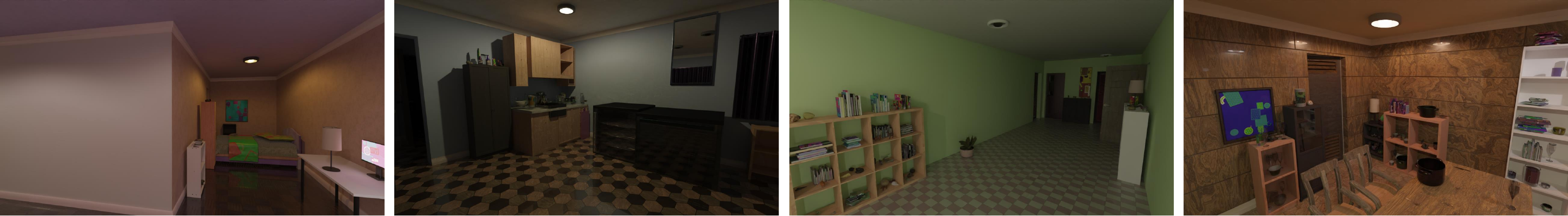}}

    \subfloat[\textit{BlenderKit} split\label{fig:qualitative_blenderkit}]
    {\includegraphics[width=\linewidth]{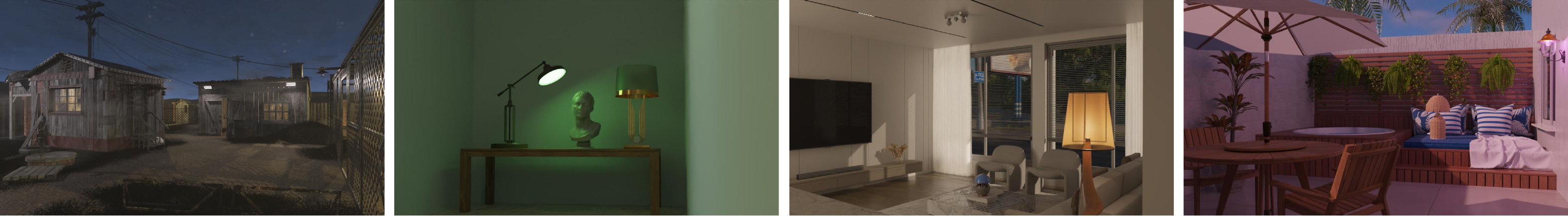}}

    \subfloat[\textit{Real} split\label{fig:qualitative_realsplit}]
    {\includegraphics[width=\linewidth]{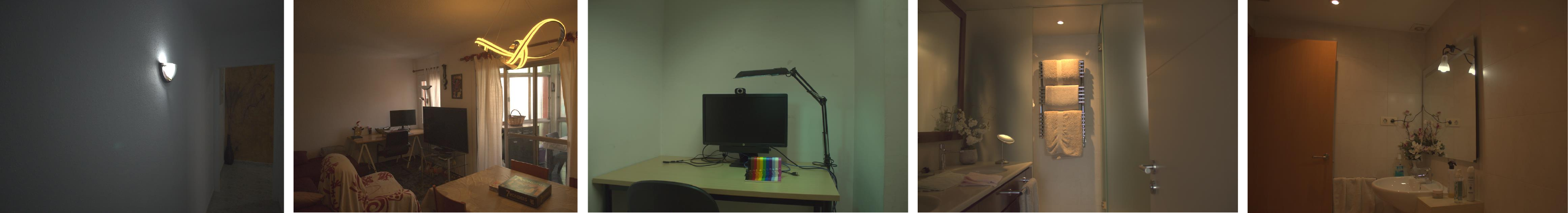}}
    \vspace{-2mm}
    \caption{Examples from the SyncLight dataset. Samples were selected randomly with random light colors.}
    \label{fig:dataset_examples}
\end{figure*}

\Cref{fig:dataset_examples} illustrates examples from the SyncLight dataset. Images and light conditions were selected randomly.

\section{Further method analysis and details}
\paragraph{Interpolant stochasticity.} Flow matching interpolates between two samples through a deterministic linear combination. In contrast, bridge matching introduces a stochastic component via Gaussian noise:

\begin{equation}
\mathbf{z}_t = (1 - t)\mathbf{z}_{src} + t\mathbf{z}_{tar} + \sigma \sqrt{t(1-t)} \boldsymbol{\epsilon} \,,\label{eq:supp_interpolant}
\end{equation}

where $\mathbf{z}_{\text{src}}$ and $\mathbf{z}_{\text{tar}}$ are (latent) samples from the source and target distributions, respectively, and $\boldsymbol{\epsilon} \sim \mathcal{N}(0, \mathbf{I})$. The parameter $\sigma$ controls the level of stochasticity in the interpolant. \Cref{tab:sigma} evaluates four different values of $\sigma$. For each row, we retrain SyncLight using four steps ($T=4$), but with different values of $\sigma$. We find that $\sigma=0.005$ achieves optimal performance, while both deterministic flow matching ($\sigma=0$) and higher noise levels ($\sigma>0.005$) result in degraded performance.

\paragraph{Number of timesteps $T$ during training.} We employ four timesteps during training. It is crucial to learn a well-conditioned vector field rather than a brittle point-to-point mapping. As the broader flow matching literature shows \citep{flowmatching}, training over multiple timesteps yields a generalizable transport, even if inference is truncated to fewer steps~\citep{chadebec2025lbm}. Reducing the number of timesteps to one ($T_{train}=1$) during training reduces the formulation to an $L_2$ loss. To validate this, we trained SyncLight using only one timestep during training (using $t=0$ in \cref{eq:supp_interpolant}). \Cref{tab:trainingT} reports a 4.82dB drop in PSNR on the Real split of our dataset compared to our $T_{train}=4$ formulation. Multi-step training provides a clear, measurable benefit.

\paragraph{Number of timesteps at inference time.} SyncLight is trained with four equally spaced timesteps ($t$ in \cref{eq:supp_interpolant}). Surprisingly, we find that using only a single timestep during inference is both faster and more accurate, as shown in \cref{tab:steps}. Note that for this experiment, the model weights are the ones obtained using four timesteps during training, and even with that, one-step inference gives better performance. We attribute this counterintuitive result to the initialization strategy. Unlike standard text-to-image models that start from Gaussian noise, our framework begins with the input images as $\mathbf{z}_{\text{src}}$, which already provides a strong initialization. Under this regime, each additional timestep introduces a discretization error that accumulates across the trajectory, ultimately degrading the quality of the final output.

\paragraph{Velocity estimator discussion.} We employ Stable Diffusion as the velocity estimator in our bridge matching formulation. While video diffusion models may appear suitable for multi-image processing, their architectures are designed for temporally coherent sequences where consecutive frames exhibit high visual similarity. Our multi-view scenario differs fundamentally: synchronized views from different camera positions contain substantially different visual content despite depicting the same scene. The temporal-encoding mechanisms in video VAEs, optimized for frame-to-frame coherence, struggle to capture this sparse multi-view setting. Instead, we leverage image diffusion models and introduce explicit multi-view reasoning through our transformer-based attention mechanism. \rev{We also experimented with Flux-based DiT backbones early on, but they failed to propagate the relighting to the additional views.}

\paragraph{Multi-view transformer block further details.} SyncLight takes as input two synchronized, uncalibrated views under source lighting, together with a target lighting condition specified on the reference view, and predicts both views under the target lighting. Standard diffusion- and flow-based architectures are designed for single-image inputs and do not model cross-view interactions. We therefore modify the transformer blocks of Stable Diffusion XL to operate in our setting while preserving its pretrained weights. In contrast to text-to-3D methods such as MVDream~\citep{mvdream} and Wonder3D~\citep{long2024wonder3d}, our approach does not require camera poses, as we aim at synchronized light source editing rather than 3D generation. The key idea is to enable information exchange across $N$ views through a modified self-attention mechanism (see~\cref{fig:multi-view transformer_block} for an illustration with $N=2$). Given a batch of $N$ view latents, we first process each view's features independently through the initial convolutional layers. Before each transformer block, features from all views are concatenated along the token dimension rather than the batch dimension, allowing features from different views to interact within the multi-view transformer block. \Cref{fig:interview2} shows an example with $N=3$ views, where a table lamp is turned on with a red color and the angle between the reference view and one of the other views reaches up to $90^\circ$

\paragraph{Lightmap codification.} Finally, we analyze the choice of color space for lightmap encoding. We employ the CIE Lab color space to encode target light colors, as it provides a perceptually uniform color representation. \Cref{tab:colorspace} compares Lab encoding against RGB. While performance is similar, Lab achieves marginally better results and offers user-friendliness. Its inherent separation of luminance (L) and chromaticity (ab) enables independent control over light intensity and color, which is more intuitive for users.

\begin{figure*}[t!]
    \centering
    \includegraphics[width=\linewidth]{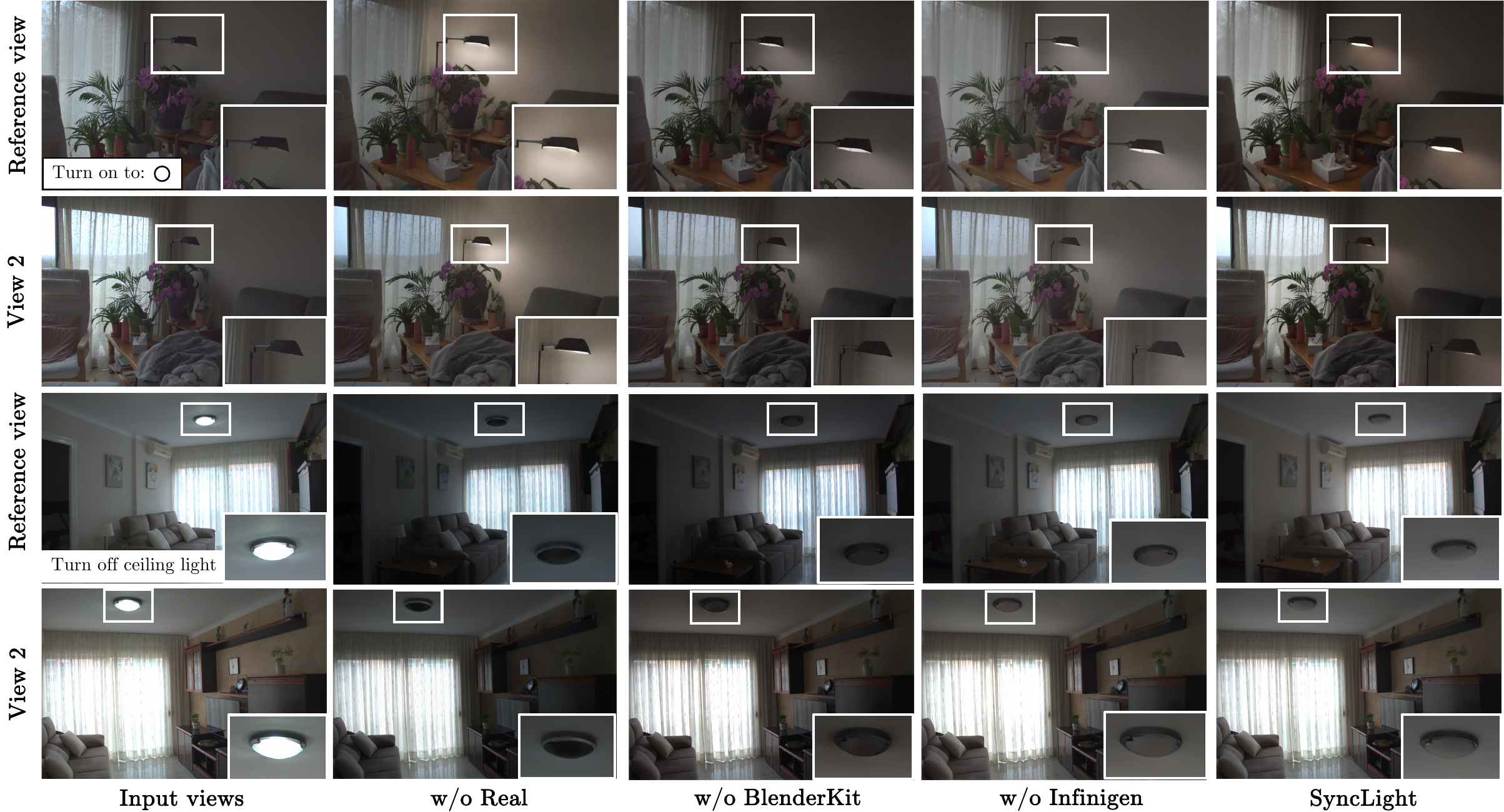}
    %\vspace{-5mm}
    \caption{Multi-view relighting examples when training SyncLight without one of the splits of the dataset at a time.}
  \label{fig:ablation_images}
\end{figure*}

\paragraph{The effect of diverse domains} \Cref{fig:ablation_images} complements the Ablation study from the main paper, showing qualitative results from our model when each of the dataset splits is removed at a time. In detail, we can see how lamps can lose realism when either the \textit{Real captures} or the \textit{BlenderKit} splits are not used—see the dark color of the ceiling lamp in the bottom figure. In contrast, not using the large number of images from the \textit{Infinigen} split forces the model to output slightly brighter results.

\paragraph{Implementation details.}
We fine-tune our modified SDXL model \cite{podell2024sdxl} for 100,000 steps using a learning rate of $1\times10^{-5}$ and a batch size of 6 at a resolution of $1280 \times 720$. Training is performed on six NVIDIA A40 GPUs \rev{and takes 8 days}. $\lambda=10$ in all experiments. \rev{Although \emph{Infinigen} dominates the pair count ($\sim$920K of $\sim$985K pairs), during training we sample uniformly from the three splits, so the \emph{Real} and \emph{BlenderKit} splits are effectively upweighted. The splits play complementary rather than interchangeable roles: \emph{Infinigen} provides cheap geometric diversity that helps preserve the pretrained prior, while the other two provide realism (\cref{tab:ablations}).} % CHECK: the rebuttal to JUD6 says "4 NVIDIA A40 GPUs for 8 days", but this paragraph says six A40s. Make both consistent.

\begin{table}
\begin{minipage}[t!]{0.48\linewidth}
    \centering
    \caption{Effect of the noise level $\sigma$.}
    \vspace{2mm}
    \setlength{\tabcolsep}{4pt}
    \label{tab:sigma}
    \begin{tabular}{lcccc}
      \toprule
      & PSNR & SSIM & $\Delta E_{00}$ & LPIPS \\
      \midrule
      $\sigma = 0$ (FM)  & 29.71 & .883 & 3.91 & .203 \\
      $\sigma = 0.001$   & 30.03 & .884 & 3.87 & .210 \\
      $\sigma = 0.005$   & 30.29 & .896 & 3.65 & .196 \\
      $\sigma = 0.01$    & 26.48 & .863 & 4.76 & .228 \\
      \bottomrule
    \end{tabular}
  \end{minipage}
  \hfill
    \begin{minipage}[t!]{0.48\linewidth}
    \centering
    \caption{Effect of the number of steps during training $T$. For inference, we use only one timestep on both cases.}
    \vspace{2mm}
    \setlength{\tabcolsep}{4pt}
    \label{tab:trainingT}
    \begin{tabular}{lcccc}
      \toprule
      & PSNR & SSIM & $\Delta E_{00}$ & LPIPS \\
      \midrule
      $T_{train} = 1$ & 25.47 & .858 & 5.02 & .243 \\
      $T_{train} = 4$   & 30.29 & .896 & 3.65 & .196 \\
      \bottomrule
    \end{tabular}
  \end{minipage}
\end{table}

\begin{figure}
    \begin{minipage}[t!]{0.48\linewidth}
    \centering
    \setlength{\tabcolsep}{4pt}
    \captionof{table}{Effect of the number of inference steps.}
    \label{tab:steps}
    \begin{tabular}{lcccc}
      \toprule
      & PSNR & SSIM & $\Delta E_{00}$ & LPIPS \\
      \midrule
      One-step   & 30.29 & .896 & 3.65 & .196 \\
      Two-step   & 30.04 & .893 & 3.71 & .197 \\
      Four-step  & 29.98 & .893 & 3.74 & .200 \\
      \bottomrule
    \end{tabular}
  \end{minipage}
  \hfill
  \begin{minipage}[t!]{0.48\linewidth}
    \centering
    \captionof{table}{Effect of the color space used for the LBM formulation.}     \label{tab:colorspace}
    \setlength{\tabcolsep}{4pt}
    \begin{tabular}{lcccc}
      \toprule
      & PSNR & SSIM & $\Delta E_{00}$ & LPIPS \\
      \midrule
      RGB     & 30.09 & .894 & 3.82 & .199 \\
      CIE Lab & 30.29 & .896 & 3.65 & .196 \\
      \bottomrule
    \end{tabular}
  \end{minipage}  
\end{figure}

\begin{table}
\begin{minipage}[t]{0.50\linewidth}
    \centering
    \includegraphics[width=\linewidth]{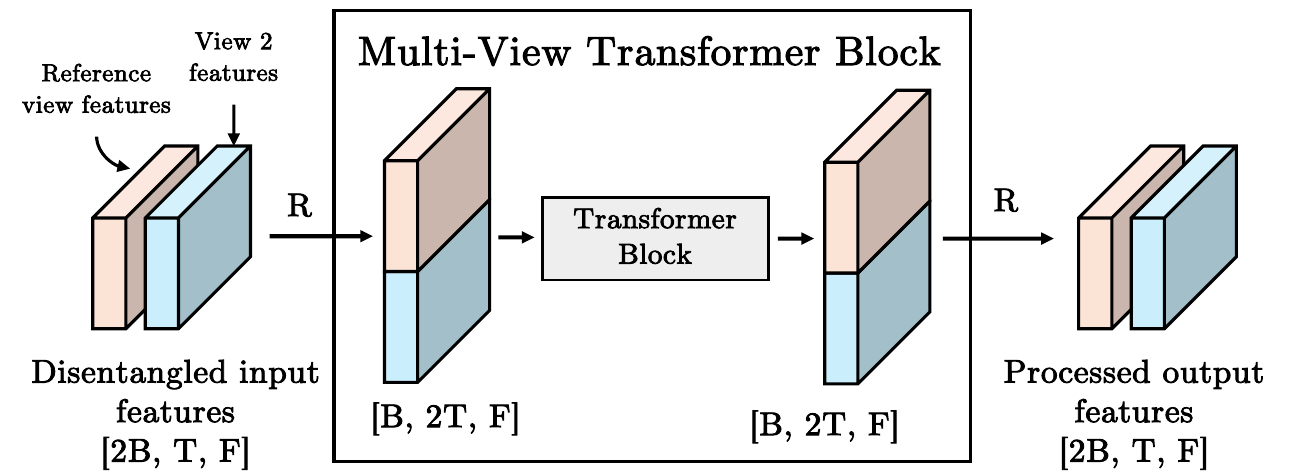}
    \vspace{2mm}
    \captionof{figure}{Multi-view transformer block. To ensure consistent relighting across viewpoints, we modify the standard self-attention mechanism of SDXL. View features are concatenated and reshaped along the sequence dimension, creating a unified representation of shape $[B, 2T , F]$}
    \label{fig:multi-view transformer_block}
  \end{minipage}
  \hfill
    \begin{minipage}[t]{0.46\linewidth}
    \centering
    \includegraphics[width=\linewidth]{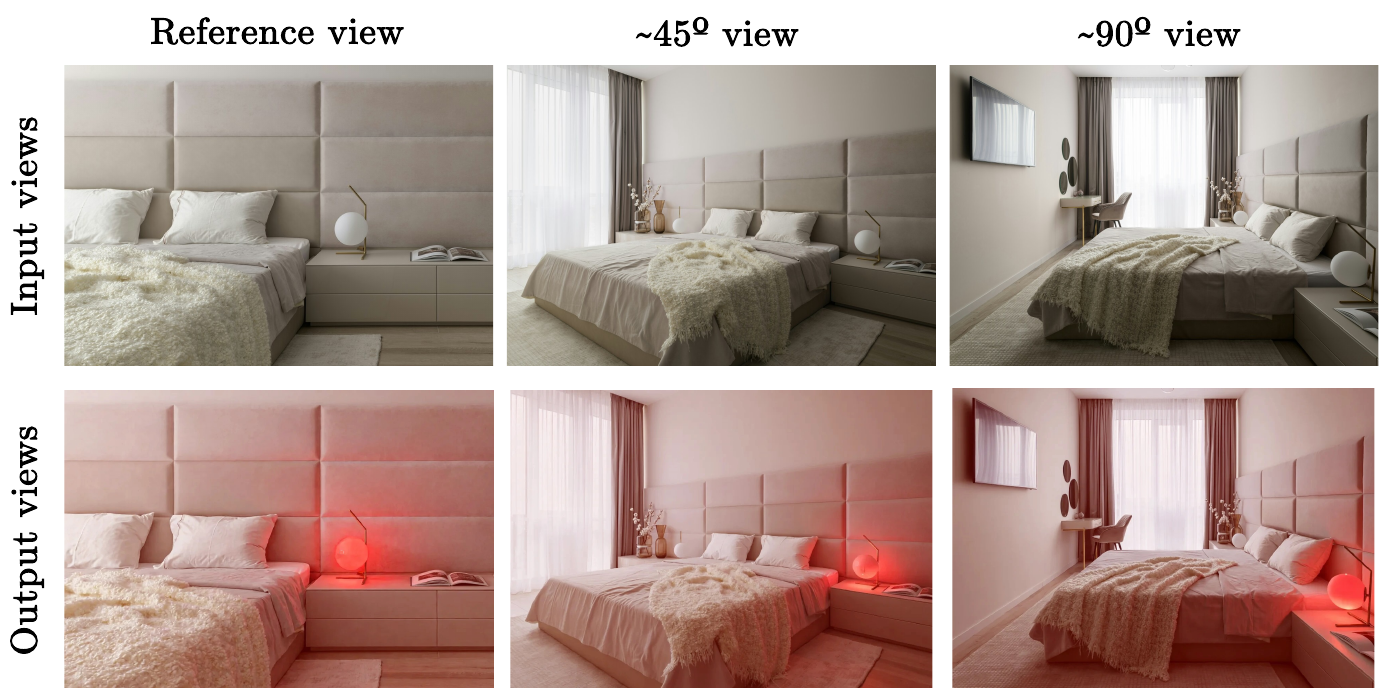}
    \vspace{2mm}
    \captionof{figure}{Example of SyncLight with $N=3$ views. Despite up to $\sim\!90^\circ$ between the reference view and the farthest view, SyncLight produces consistent relighting across all viewpoints when turning on a red table lamp.}
    \label{fig:interview2}
  \end{minipage}  
\end{table}

% \begin{figure*}[t!]
%     \centering
%     \includegraphics[width=\linewidth]{figures/supp_ablation.pdf}
%     %\vspace{-5mm}
%     \caption{Multi-view relighting examples when training SyncLight without one of the splits of the dataset at a time.}
%   \label{fig:ablation_images}
% \end{figure*}

\begin{figure}[t!]
 %   \centering
    \includegraphics[width=0.98\linewidth]{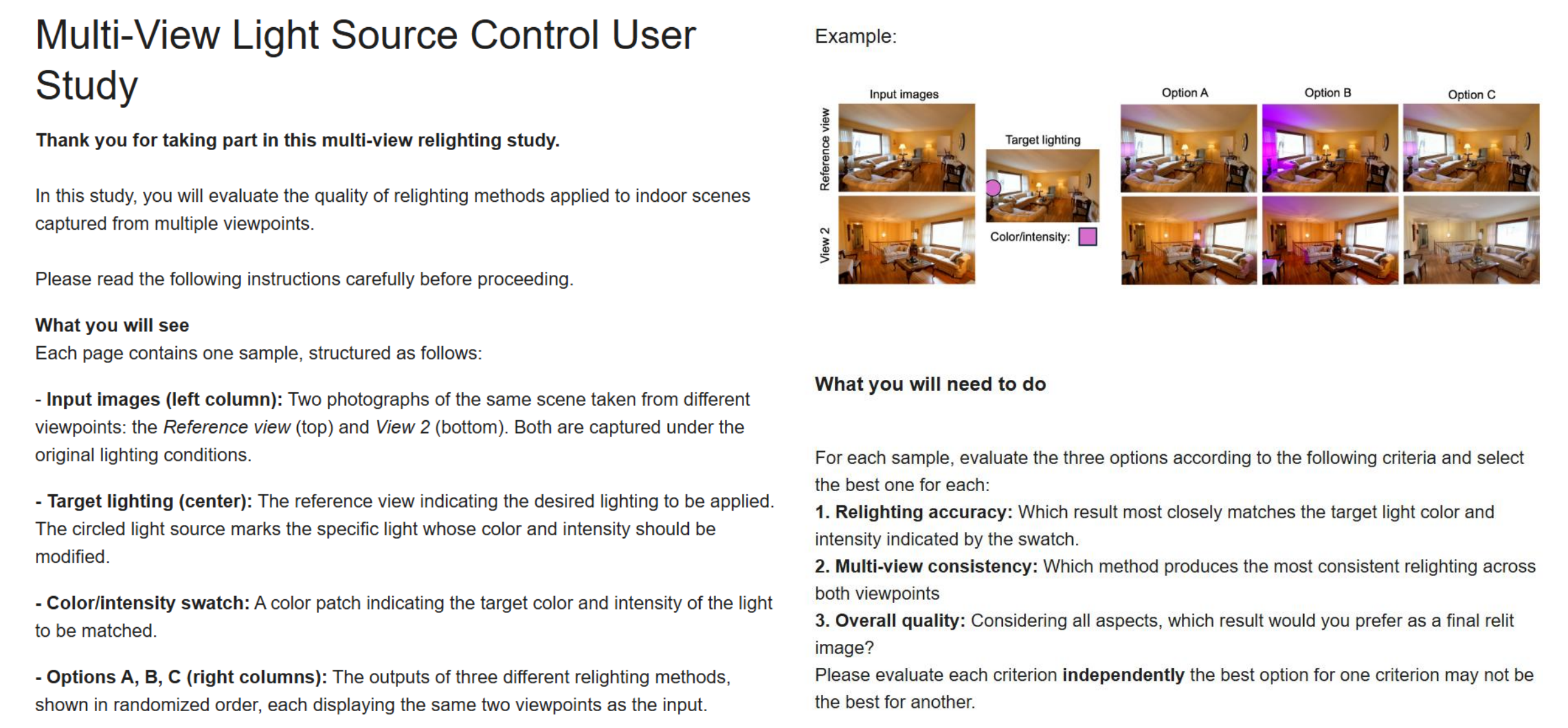}
    %\vspace{-5mm}
    \caption{Instructions given to the users for the User Study.}
  \label{fig:instructions}
\end{figure}

\begin{figure}[t!]
    \centering
    \includegraphics[width=0.6\linewidth]{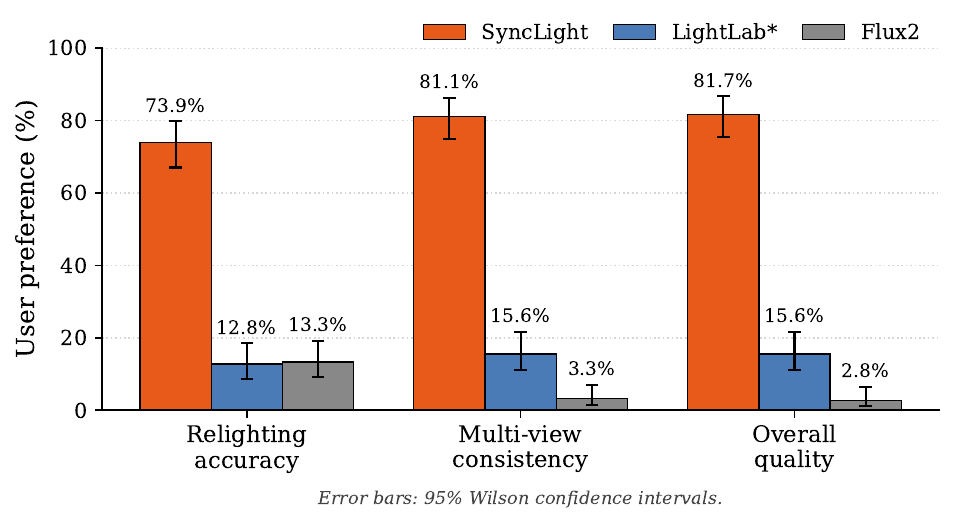}
    %\vspace{-5mm}
    \caption{User study results.}
  \label{fig:user_study_results}
\end{figure}

\begin{figure*}[t!]
    \centering
    \subfloat[Video relighting. From user inputs in the first frame only, SyncLight automatically relights 142 frames of a video from the RealEstate10k dataset~\cite{RealEstate10K}. \label{fig:video_application}]
    {\includegraphics[width=\linewidth]{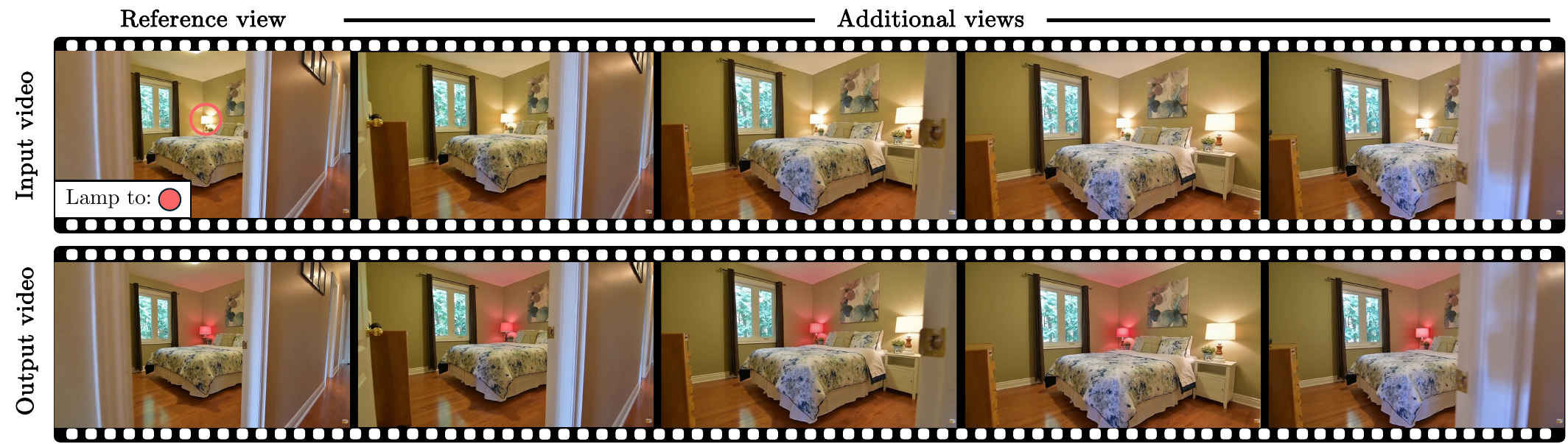}}
    \vspace{-0.5em}

    \subfloat[Novel-view synthesis. 3DGS~\cite{gaussian_splatting} is applied to images relit by SyncLight, which achieves out-of-the-box multi-view consistency. \label{fig:3dgs_application}]
    {\includegraphics[width=\linewidth]{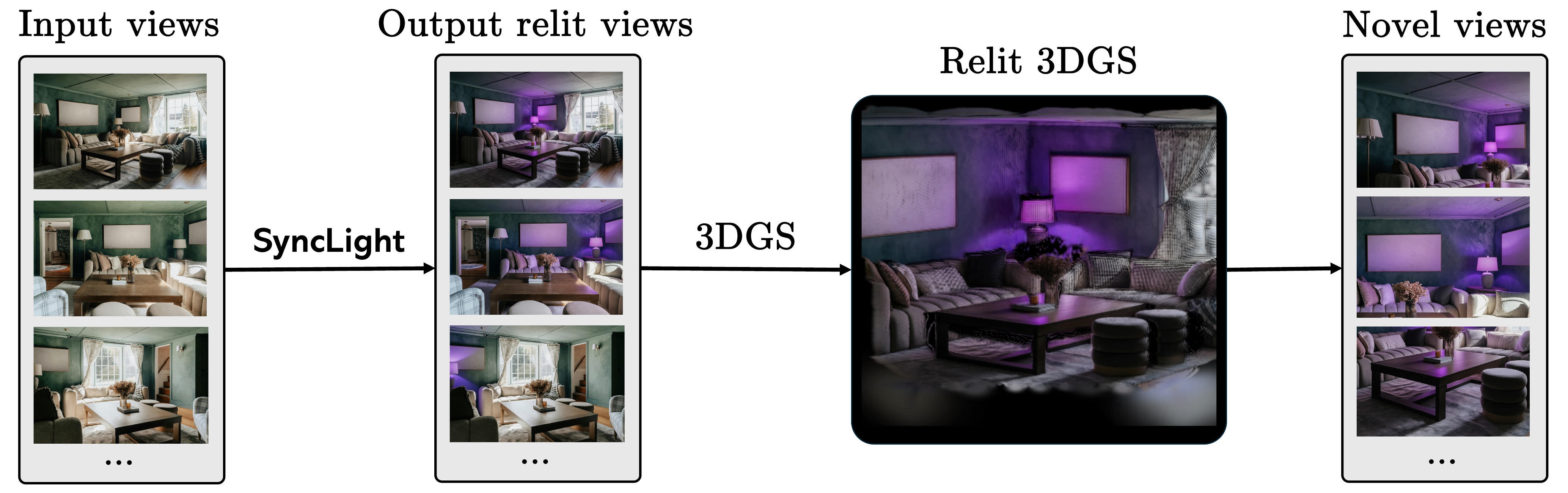}}
    
    \caption{SyncLight applied to: (a) video relighting, and (b) novel-view synthesis with radiance fields. Please see the supplementary video on our webpage.}
    \label{fig:applications_synclight}
\end{figure*}

\begin{figure*}[t!]
    \centering
    %\subfloat[\textit{Infinigen} split\label{fig:qualitative_infinigen}]
    {\includegraphics[width=\linewidth]{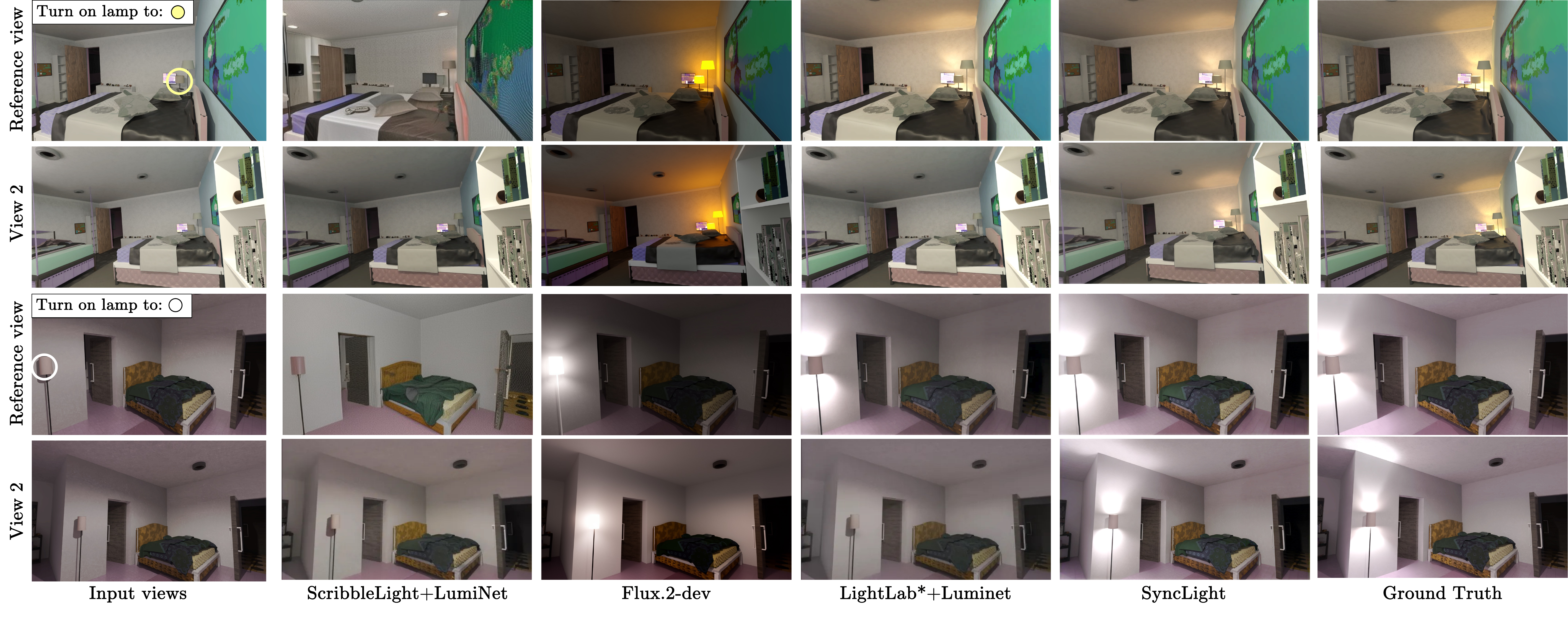}}

    \vspace{2mm}

    %\subfloat[\textit{BlenderKit} split\label{fig:qualitative_infinigen}]
    {\includegraphics[width=\linewidth]{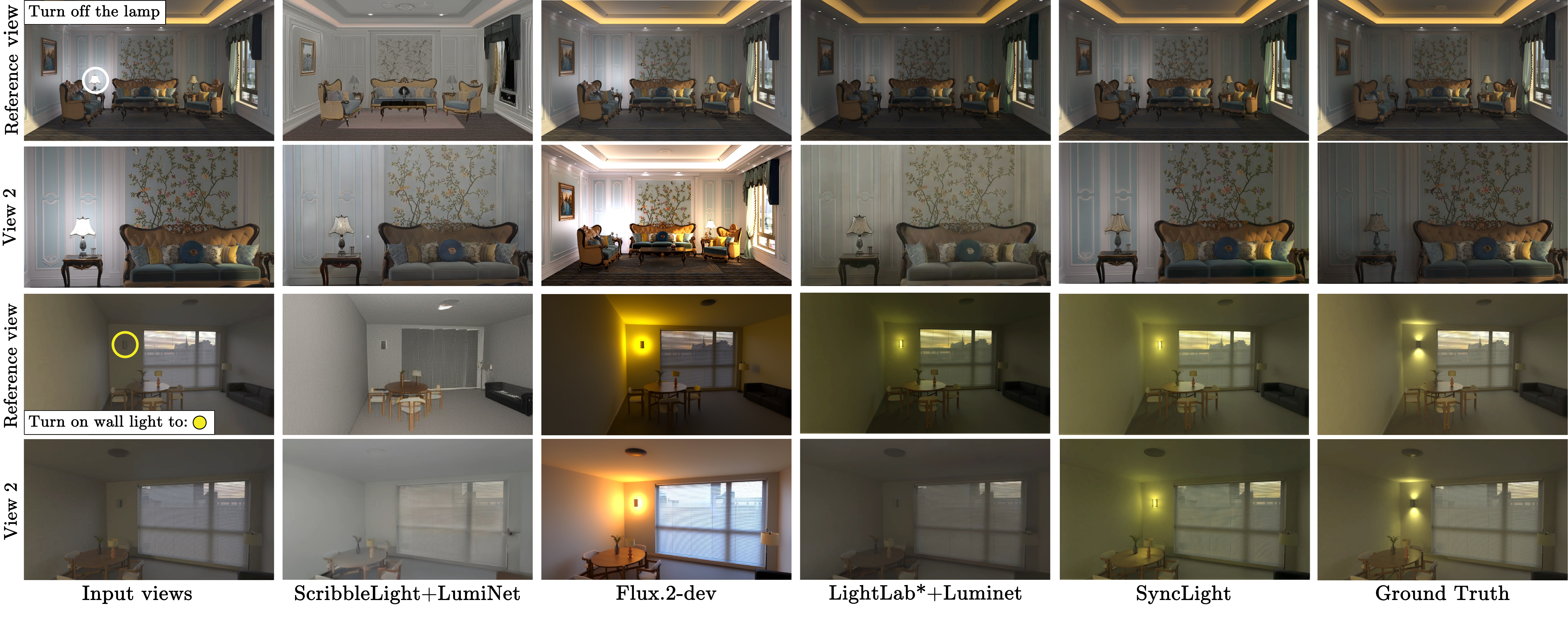}}

    \vspace{-2mm}
    \caption{Qualitative results on the SyncLight dataset on the  \emph{Infinigen} (1--2), and \emph{BlenderKit} (3--4) splits. \emph{Real} split is shown in the main paper.}
    \label{fig:qualitative_synclight}
\end{figure*}

\section{\texorpdfstring{\rev{Extensions and future directions}}{Extensions and future directions}}
\label{sec:supp_extensions}
{\revcolor
Each ingredient of SyncLight generalizes independently. \emph{Outdoor scenes} could be handled by conditioning on sun direction and strength~\citep{holdgeoffroy2017deep}, on encoded environment maps~\citep{he2025unirelight}, or on a lighting token~\citep{zhang2026unilight}. \emph{Windows and area lights} only require replacing circular markers with segmentation masks in the same 4-channel Lab encoding, a drop-in change to the conditioning rather than to the architecture. \emph{Dynamic scenes} need no architectural change either: our multi-view attention treats its input as a set, so the $N\times T$ latents of a multi-camera video can be processed as one set of tokens while the user still edits a single frame; training only requires adding temporal pairs to the batch. Finally, an \emph{agentic interface} could remove manual interaction: a VLM parses a text prompt, a segmentation model such as SAM~\citep{kirillov2023sam} localizes the named source in the reference view, and pseudo-Lab values are derived from the color name, making text and clicks two entry points to the same conditioning, with text trading some color and intensity precision for convenience.
}

\section{\texorpdfstring{\rev{Cross-view consistency metric (VGGTm)}}{Cross-view consistency metric (VGGTm)}}
\label{sec:supp_vggtm}
{\revcolor
Per-image metrics (PSNR, SSIM, $\Delta E_{00}$, LPIPS) compare each output with its own ground truth and therefore cannot tell whether two outputs agree with each other: a model can score well on every view while producing geometrically incompatible views. VGGTm addresses this with VGGT~\citep{wang2025vggt}. For each test scene, we extract correspondences between the two ground-truth views and, separately, between the two output views, and count the high-confidence matches in each case. VGGTm is the ratio between the output count and the ground-truth count, expressed as a percentage. A ratio near 100\% indicates that the relit views remain as mutually 3D-consistent as the ground truth, whereas a lower ratio reveals cross-view hallucinations, such as shadows or highlights that do not agree across views, that pixel-wise metrics miss.

Because VGGT's correspondence confidence is scene-dependent (it degrades on wide-baseline or low-overlap pairs regardless of relighting quality), we gate applicability on the ground-truth pair's own confidence, which is independent of the model under evaluation, and report the metric only on scenes that clear this bar. For additional views, the metric is averaged over view pairs.
}
% CHECK: add the confidence threshold, the fraction of scenes that pass the gate, and how pairs are formed for the additional views.

\section{\texorpdfstring{\rev{Scalability analysis}}{Scalability analysis}}
\label{sec:supp_scaling}
{\revcolor
\Cref{fig:scaling_plot} plots inference time and peak memory against the number of views $N$ for the 32-view BlenderKit experiment of \cref{tab:scaling} in the main paper. We use memory-efficient attention (FlashAttention~\citep{dao2022flashattention} through PyTorch SDPA), so the $NT \times NT$ attention matrix is never materialized and memory scales as $\mathcal{O}(N\cdot T)$: it grows linearly, from 15.4\,GB at $N=2$ to 20.9\,GB at $N=32$. Compute grows quadratically with $N$, as expected for full cross-view attention, yet remains practical. \Cref{tab:scaling_real} repeats the analysis on the Real test split for $N\in\{2,4,6,7\}$: performance is consistent regardless of the number of views. Beyond these ranges, the applications in \cref{sec:supp_applications} push $N$ much further: a 142-frame video is relit jointly in 156\,s, and radiance-field reconstruction from relit views remains consistent.
}

\begin{figure}[t]
  {\revcolor
  \centering
  \begin{tikzpicture}
    \begin{axis}[width=0.48\linewidth, height=4.2cm, xmode=log, log basis x=2,
      xtick={2,4,8,16,32}, xticklabels={2,4,8,16,32},
      xlabel={Number of views $N$}, ylabel={Inference time (s)},
      ymin=0, grid=major, tick label style={font=\footnotesize}, label style={font=\footnotesize}]
      \addplot[thick, mark=*, color=blue!70!black] coordinates {(2,1.58) (4,2.17) (8,5.82) (16,17.39) (32,43.71)};
    \end{axis}
  \end{tikzpicture}\hfill
  \begin{tikzpicture}
    \begin{axis}[width=0.48\linewidth, height=4.2cm, xmode=log, log basis x=2,
      xtick={2,4,8,16,32}, xticklabels={2,4,8,16,32},
      xlabel={Number of views $N$}, ylabel={Peak VRAM (GB)},
      ymin=10, ymax=24, grid=major, tick label style={font=\footnotesize}, label style={font=\footnotesize}]
      \addplot[thick, mark=square*, color=orange!80!black] coordinates {(2,15.4) (4,15.8) (8,16.9) (16,18.3) (32,20.9)};
    \end{axis}
  \end{tikzpicture}
  \vspace{-2mm}
  \caption{\rev{Inference time (left) and peak memory (right) of SyncLight as a function of the number of views $N$ on an NVIDIA A40 (BlenderKit, 32 rendered views per scene). Memory grows linearly; compute grows quadratically.}}
  \label{fig:scaling_plot}
  }
\end{figure}

\begin{table}[t]
  \centering
  {\revcolor
  \caption{\rev{Scaling with the number of views $N$ on the Real test split.}}
  \label{tab:scaling_real}
  \setlength{\tabcolsep}{4pt}
  \begin{tabular}{lcccccc}
    \toprule
    $N$ & PSNR & SSIM & $\Delta E_{00}$ & LPIPS & VGGTm & Inf. time (s) \\
    \midrule
    2 & 30.23 & .897 & 3.56 & .197 & 95.8 & 1.58 $\pm$ 0.02 \\
    4 & 30.40 & .904 & 3.55 & .194 & 94.8 & 2.17 $\pm$ 0.02 \\
    6 & 30.45 & .907 & 3.56 & .194 & 94.9 & 3.01 $\pm$ 0.03 \\
    7 & 30.47 & .906 & 3.56 & .192 & 95.3 & 3.83 $\pm$ 0.03 \\
    \bottomrule
  \end{tabular}
  }
\end{table}

\begin{table}[t]
  {\revcolor
  \centering
  \caption{\rev{User input required by each method.}}
  \label{tab:interaction}
  \small
  \begin{tabular}{lp{0.37\linewidth}p{0.37\linewidth}}
    \toprule
    Method & Interaction per edited light & Cost per additional view \\
    \midrule
    SyncLight & one click + color + intensity & none \\
    LightLab~\citep{lightlab} & segmentation mask + color + intensity & one mask per view \\
    GR3EN~\citep{xiaoyan2026gr3en} & segmentation mask + color + intensity & one mask per view + camera poses \\
    LuxRemix~\citep{ruofan2026luxremix} & OLAT harmonization + color + intensity & Pl\"ucker embeddings + harmonization \\
    \bottomrule
  \end{tabular}
  }
\end{table}

\section{\texorpdfstring{\rev{Local versus global illumination changes}}{Local versus global illumination changes}}
\label{sec:supp_ambience}
{\revcolor
Editing a light source can change the overall ambience of the scene. This is physically expected rather than a degradation: toggling or recoloring a light changes the indirect illumination throughout the scene, and the ground truth exhibits the same shift. Our targets are formed by OLAT composition in linear space (HDR renders, camera RAW), $I_\text{relit} = I_\text{amb} + \sum_i c_i L_i$, the same procedure used by LightLab~\citep{lightlab}, LuxRemix~\citep{ruofan2026luxremix}, and GR3EN~\citep{xiaoyan2026gr3en}, with roots in light stage capture~\citep{debevec2000acquiring}. Because the combination is linear, the ground truth is photometrically exact: the ambient term $I_\text{amb}$ is preserved, and the relative size of the ambience shift is set by the ratio between the edited source and the illumination already present. A scene lit by strong daylight therefore changes little when a dim source is turned on.

To verify that SyncLight does not shift the ambience more than the ground truth, \cref{tab:ambience} separates weak from strong edits, measured by the change in requested lightness $|\Delta L|$, and groups pixels into rings around the edited source at increasing distance, in multiples of the marker radius $r$. The first two metric columns compare our output with the ground truth; the last two compare the ground truth and our output with the input image, measuring how much the scene changed in the first place. For weak edits, pixels far from the edited light stay close to the input: the ground truth changes by 0.9\,$\Delta E_{00}$ there and our output by 1.0, so the ambience is nearly unchanged and SyncLight does not shift it more than the ground truth does. For strong edits, both numbers are much larger, as the far region covers most of the image and absorbs the indirect illumination. In both cases, our output tracks the ground truth, so the size of the ambience shift is set by the scene and the edit, not by the model.
}

\begin{table}[t]
  \centering
  {\revcolor
  \caption{\rev{Local versus global changes on the Real split. Weak and strong edits are separated by the requested lightness change $|\Delta L|$; pixels are grouped by their distance to the edited source in multiples of the marker radius $r$. The last row reports all edits over the full image, as in \cref{tab:quantitative}.}}
  \label{tab:ambience}
  \small
  \setlength{\tabcolsep}{5pt}
  \begin{tabular}{lcccc}
    \toprule
    & \multicolumn{2}{c}{Ours vs.\ GT} & \multicolumn{2}{c}{$\Delta E_{00}$ vs.\ input} \\
    \cmidrule(lr){2-3}\cmidrule(lr){4-5}
    Region & PSNR & $\Delta E_{00}$ & GT & Ours \\
    \midrule
    \multicolumn{5}{l}{\emph{Weak edits} ($|\Delta L| < 0.5$)} \\
    \quad 0--1\,$r$ (marker) & 30.6 & 3.12 & 7.9 & 8.2 \\
    \quad 1--2\,$r$ & 30.9 & 2.84 & 3.1 & 3.3 \\
    \quad 2--4\,$r$ & 31.8 & 2.62 & 1.6 & 1.7 \\
    \quad $>$4\,$r$ & 32.1 & 2.71 & 0.9 & 1.0 \\
    \quad Full image & 31.7 & 2.83 & 1.4 & 1.5 \\
    \midrule
    \multicolumn{5}{l}{\emph{Strong edits} ($|\Delta L| > 0.5$)} \\
    \quad 0--1\,$r$ (marker) & 33.1 & 6.31 & 23.4 & 23.9 \\
    \quad 1--2\,$r$ & 32.8 & 4.78 & 12.7 & 13.1 \\
    \quad 2--4\,$r$ & 31.0 & 4.26 & 7.4 & 7.6 \\
    \quad $>$4\,$r$ & 29.2 & 4.08 & 5.8 & 6.0 \\
    \quad Full image & 29.9 & 4.21 & 7.1 & 7.3 \\
    \midrule
    All edits, full image & 30.2 & 3.56 & -- & -- \\
    \bottomrule
  \end{tabular}
  }
\end{table}

\section{User study}
Standard image-quality metrics such as PSNR, SSIM, $\Delta E_{00}$, and LPIPS measure pixel- or feature-level distance to a reference image, but do not capture key perceptual aspects of relighting: fidelity to the requested light colour and intensity, plausibility of shadows and inter-reflections, and coherence across viewpoints. We therefore complement our quantitative evaluation with a user study assessing these perceptual qualities directly.

We present the instructions given to participants in \cref{fig:instructions}. Participants were first introduced to the overall layout of the experimental interface through both written instructions and a visual example, and were then given explanations of the three types of questions they would be asked. For each sample, participants were shown a reference view together with a target lighting specification and the corresponding outputs from three methods: SyncLight, LightLab* (provided with a per-view lightmap)~\citep{lightlab}, and Flux2~\citep{flux2_2025}. Although our method internally uses CIE Lab, we present the target lighting to participants as an RGB swatch encoding the desired color and intensity, for simplicity. The light source to be manipulated is indicated by a circle marker overlaid on the reference view. The order of the three methods was randomized per sample to eliminate position bias. Participants were asked to answer three questions, each targeting a distinct perceptual axis of relighting quality:

\begin{enumerate}[leftmargin=*,noitemsep]
\item \textbf{Relighting accuracy:} Which output is closest to the target light color and intensity? This question evaluates fidelity to the user-specified edit, independent of considerations of consistency or realism.
\item \textbf{Multi-view consistency:} Which output exhibits the most consistent lighting between the two views (e.g., coherent shadow direction, matching color casts, agreement on light source state)? This question isolates the cross-view coherence that distinguishes SyncLight from per-view baselines.
\item \textbf{Overall quality:} Which output is the most visually convincing overall? This question captures the holistic perceptual judgement that integrates accuracy, consistency, and the absence of artefacts.
\end{enumerate}

We collected responses from 20 participants, each answering all three questions for 10 samples, yielding 200 votes per question. \Cref{fig:user_study_results} summarizes the results. SyncLight is preferred across all three criteria by 73.9\% (relighting accuracy), 81.1\% (multi-view consistency), and 81.7\% (overall quality), with non-overlapping 95\% Wilson confidence intervals between SyncLight and the next-best method in every case. Notably, and in line with our motivation and the results in \cref{tab:quantitative}, SyncLight outperforms LightLab* in all three criteria despite the latter being provided with a per-view lightmap as additional input. This indicates that jointly processing all views improves not only multi-view consistency but also single-view relighting accuracy, since the model can leverage cross-view evidence (e.g., shading cues visible in one view but not in another) to better estimate the target lighting in each frame. Flux2 achieves results comparable to LightLab* on relighting accuracy, which we attribute to its strength as a general-purpose image editor capable of producing plausible global colour and contrast adjustments. However, as the consistency and overall quality results show, this comes at the cost of geometric coherence across views and frequently introduces hallucinations or physically implausible artifacts.

\vspace{-5pt}
\section{Applications}
\label{sec:supp_applications}
SyncLight is trained on pairs of views $N=2$. However, in the main manuscript, we show how our approach generalizes zero-shot to multiple views $N>2$. We always aim at sparse view scenarios where the synchronous views are have a displacement between them. Having said this, there are two applications in which SyncLight is useful: video relighting and 3D relighting. See the attached video to visualize the results for both applications.

\vspace{-5pt}
\paragraph{Video relighting} SyncLight can efficiently relight videos while maintaining temporal consistency, even if it is trained only on \emph{image pairs}. Note that the user only needs to specify the desired lighting condition on the first frame of the video, and SyncLight propagates the light changes across all the following frames. \Cref{fig:video_application} shows five frames from a 142-frame video from the RealEstate10k dataset~\cite{RealEstate10K} alongside their relit versions, demonstrating consistent relighting across all frames. Notably, all 142 frames are processed jointly in a single forward pass, requiring only 156 seconds and 27 GB of VRAM.

\vspace{-5pt}
\paragraph{Novel view synthesis with radiance fields} Relighting in radiance field representations (NeRFs, or 3DGS) is inherently challenging since it typically involves complex inverse rendering (e.g., \cite{tensoir}). SyncLight's view-consistent relighting enables an alternative approach: users can relight multiple 2D views and reconstruct the 3D scene, effectively using our method as a proxy for 3D relighting. \Cref{fig:3dgs_application} and the supplementary video show an example using 3DGS \citep{gaussian_splatting} implementation from NerfStudio~\citep{tancik2023nerfstudio}. Note that, as opposed to Poirier-Ginter et al.~\citep{poirier2024diffusion}, no multi-view consistency optimization was performed: we simply used vanilla 3DGS directly on the images relit by SyncLight.

\section{Additional visual results}
~\Cref{fig:qualitative_synclight} shows further qualitative results of competing methods and SyncLight across the Infinigen and BlenderKit splits of our dataset (Real split shown in the main paper), demonstrating the strong performance of our method. In contrast, Flux.2-dev struggles with cross-view consistency, producing visually plausible but geometrically inconsistent results — note the differing lamp colors across views in BlenderKit row 2. The LightLab*+LumiNet baseline, while maintaining better consistency by explicitly relighting each view, fails to propagate lighting effects when sources are not directly visible (BlenderKit row 2, second view). 

\begin{figure*}[ht!]
    \centering
    \includegraphics[width=\linewidth]{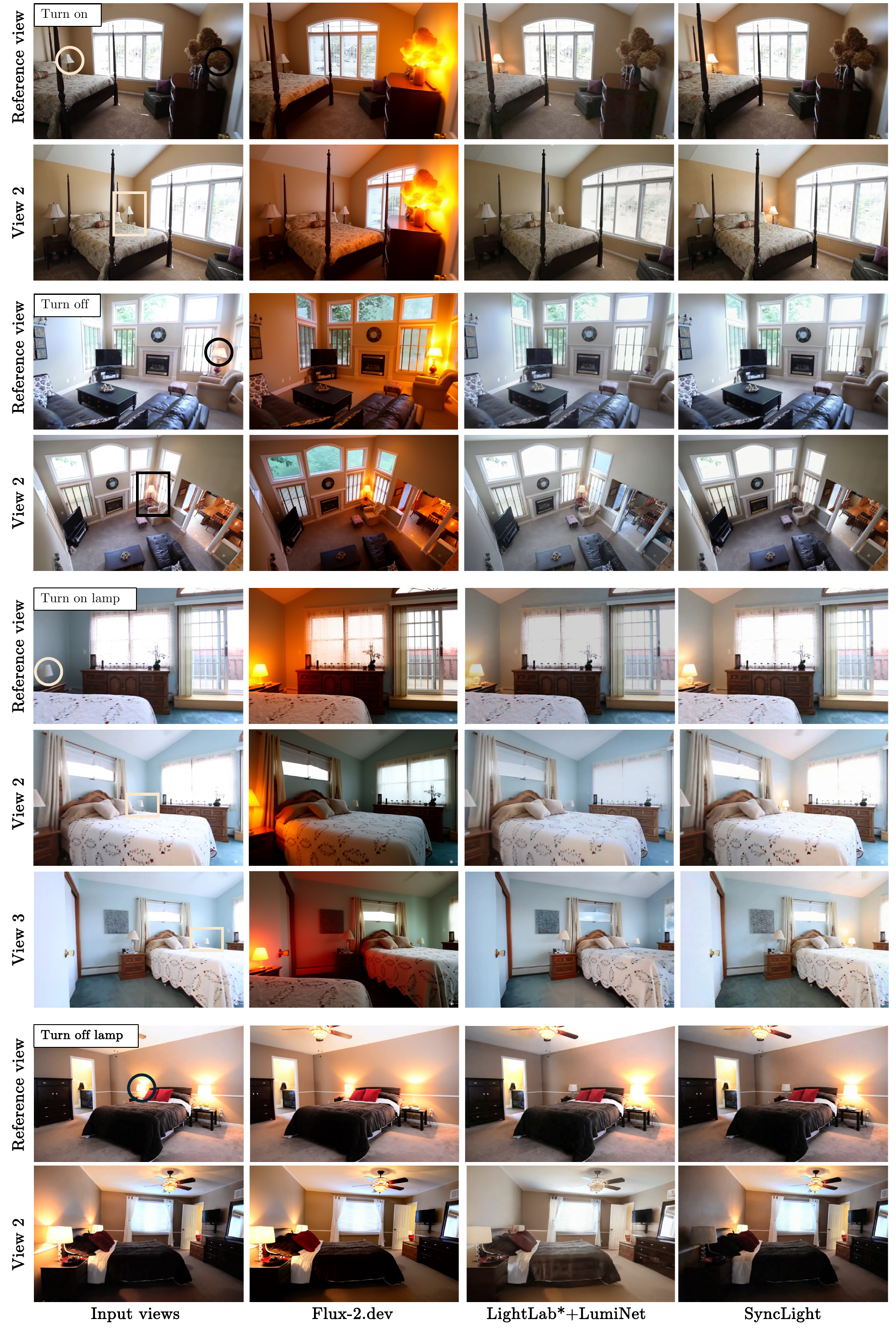}
  %  \vspace{-5mm}
\caption{Additional results on the RealEstate10K dataset.}
  \label{fig:qualitative_realstatesup2}
\end{figure*}

\begin{figure*}[ht!]
    \centering
    \includegraphics[width=\linewidth]{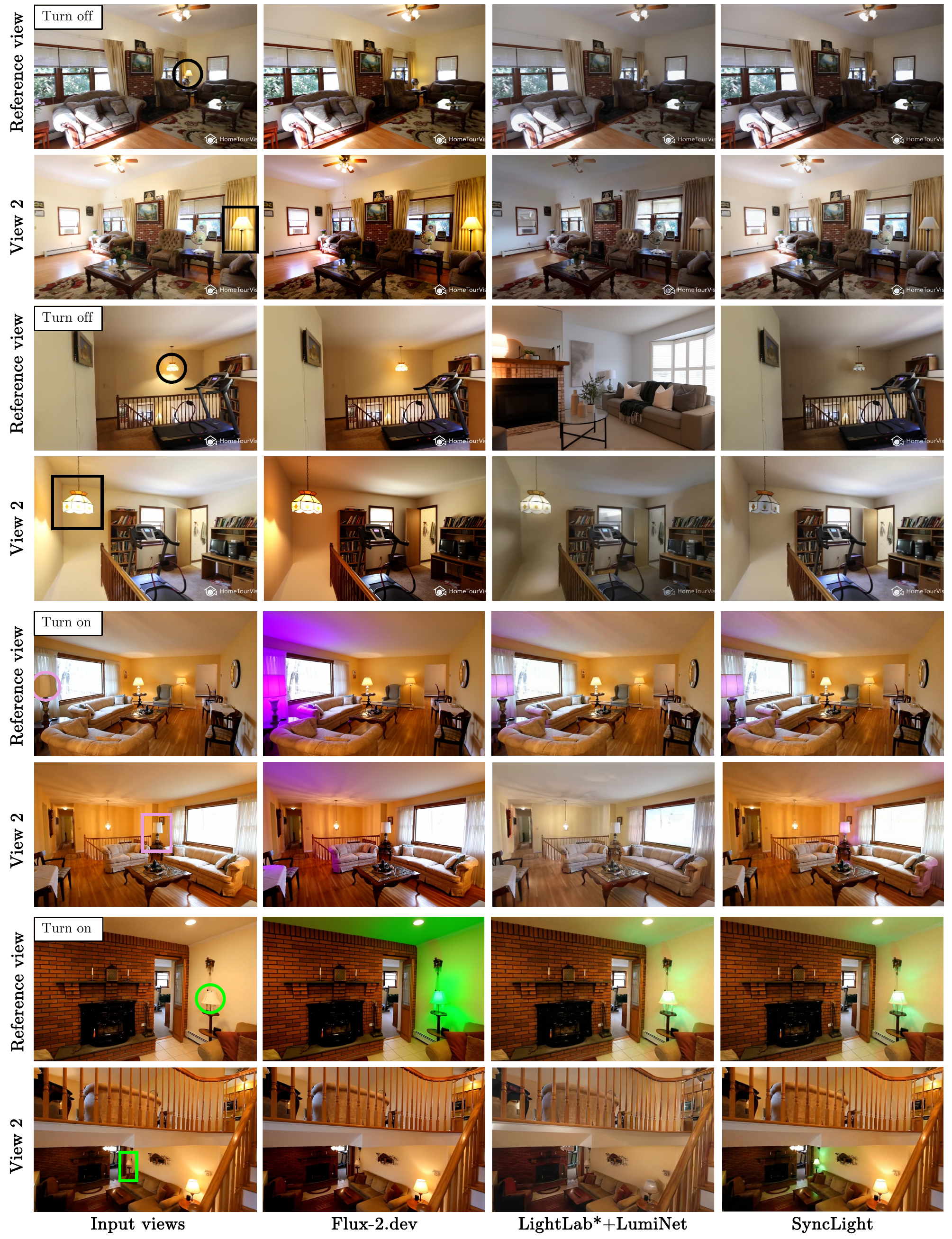}
   \vspace{-3mm}
    \caption{Additional results on the RealEstate10K dataset}
     %  \vspace{-25mm}
  \label{fig:qualitative_realstatesup3}
\end{figure*}

\Cref{fig:qualitative_realstatesup2}–\cref{fig:qualitative_realstatesup3} present additional results from our method on the RealEstate10K dataset, illustrating various scenarios in which different lights are switched on or off under different colors and intensity conditions.

% \clearpage

% \input{checklist.tex}

%%%%%%%%%%%%%%%%%%%%%%%%%%%%%%%%%%%%%%%%%%%%%%%%%%%%%%%%%%%%

\end{document}